\documentclass[review]{elsarticle}

\usepackage{hyperref}

\journal{Information Sciences}

\usepackage{xcolor}
\usepackage{amsmath}
\usepackage{amssymb}

\usepackage{algorithm}
\usepackage{algpseudocode}

\usepackage{multirow}

\newdefinition{definition}{Definition}
\newdefinition{proposition}{Proposition}
\newdefinition{lemma}{Lemma}
\newproof{proof}{Proof}

\DeclareMathOperator{\pa}{Pa}

\DeclareMathOperator{\thetab}{\boldsymbol{\theta}}

\DeclareMathOperator*{\argmax}{arg\,max}

\usepackage{array}
\newcolumntype{L}[1]{>{\raggedright\let\newline\\\arraybackslash\hspace{0pt}}m{#1}}
\newcolumntype{C}[1]{>{\centering\let\newline\\\arraybackslash\hspace{0pt}}m{#1}}
\newcolumntype{R}[1]{>{\raggedleft\let\newline\\\arraybackslash\hspace{0pt}}m{#1}}

\usepackage[justification=raggedleft]{subcaption}

\begin{document}

\begin{frontmatter}

\title{Semiparametric Bayesian Networks}

\author{David~Atienza\corref{mycorrespondingauthor}}
\ead{datienza@fi.upm.es}
\author{Concha~Bielza}
\ead{mcbielza@fi.upm.es}
\author{Pedro~Larra\~naga}
\ead{pedro.larranaga@fi.upm.es}
\address{Universidad Polit\'ecnica de Madrid, Departamento de Inteligencia Artificial, 28660 Boadilla del Monte, Spain}

\cortext[mycorrespondingauthor]{Corresponding author}

\begin{abstract}
We introduce semiparametric Bayesian networks that combine parametric and nonparametric conditional probability distributions. Their aim is to incorporate the advantages of both components: the bounded complexity of parametric models and the flexibility of nonparametric ones. We demonstrate that semiparametric Bayesian networks generalize two well-known types of Bayesian networks: Gaussian Bayesian networks and kernel density estimation Bayesian networks. For this purpose, we consider two different conditional probability distributions required in a semiparametric Bayesian network. In addition, we present modifications of two well-known algorithms (greedy hill-climbing and PC) to learn the structure of a semiparametric Bayesian network from data. To realize this, we employ a score function based on cross-validation. In addition, using a validation dataset, we apply an early-stopping criterion to avoid overfitting. To evaluate the applicability of the proposed algorithm, we conduct an exhaustive experiment on synthetic data sampled by mixing linear and nonlinear functions, multivariate normal data sampled from Gaussian Bayesian networks, real data from the UCI repository, and bearings degradation data. As a result of this experiment, we conclude that the proposed algorithm accurately learns the combination of parametric and nonparametric components, while achieving a performance comparable with those provided by state-of-the-art methods.
\end{abstract}

\begin{keyword}
Bayesian networks\sep kernel density estimation\sep semiparametric model\sep continuous data
\end{keyword}

\end{frontmatter}

\section{Introduction}

Bayesian networks~\cite{Koller} are probabilistic graphical models that are used to factorize a joint probability distribution of a set of random variables. Bayesian networks are advantageous in terms of representing the conditional independence of triplets of variables in a domain to factorize the underlying joint probability distribution. The advantages of this approach are twofold: first, the number of parameters needed to specify a model can be reduced, and the set of independencies can be easily checked by a human, providing useful knowledge about the domain of interest. Bayesian networks have been applied to solve various problems in machine learning, such as classification~\cite{BNClassifiers}, clustering~\cite{BNClustering}, and density estimation~\cite{BNAnomaly}.

Bayesian networks can be used to jointly model uncertain domains with discrete and continuous random variables. A common approach to process continuous random variables is to discretize them and learn the structure of a discrete Bayesian network from data, which can be done without assuming any underlying continuous distribution. However, this can be suboptimal due to the loss of information caused by discretization, meaning that different continuous values can be assigned to the same discrete category.

In the related literature, several methods have been introduced to model continuous random variables without discretizing data. There are two main approaches to estimate continuous probability distributions: parametric and nonparametric models. Both types of estimations have been used to construct Bayesian networks. In parametric models, a specific known probability distribution from a particular family is assumed concerning a given dataset. A distribution has a finite number of parameters that can be estimated based on data. Such models are notably efficient if the specified assumptions hold with regard to a given dataset. Nonparametric models do not assume any specific probability distribution; however, generally, an estimated probability distribution emerges from training data. Nonparametric models are more flexible, as they can represent almost any probability distribution. However, they typically have worse error convergence rates with respect to the number of instances (i.e. their estimation error decreases at a slower rate than parametric models as the sample size increases) and are associated with higher computational costs in the cases when inferences over a distribution are performed.

In the present paper, we propose to combine parametric and nonparametric estimation models to define a new kind of Bayesian networks. Hereinafter, we denote these models as semiparametric Bayesian networks (SPBNs). SPBNs are intended to combine the advantages of both parametric and nonparametric models. For this purpose, we define two types of conditional probability distributions (CPDs): parametric and nonparametric ones. A CPD assigned to each node depends on its type; this issue is detailed in Section~\ref{sec:representation}. A parametric CPD can be used to represent linear relationships between random variables using a linear Gaussian distribution. A nonparametric CPD can be considered to represent nonlinear relationships given the flexibility of nonparametric models.

The contributions of the paper are as follows: (1) definition of a new class of Bayesian networks, SPBNs, that generalize Gaussian Bayesian networks (GBNs) and kernel density estimation Bayesian networks (KDEBNs); (2) modification of the greedy hill-climbing (HC) and PC algorithms to learn SPBN structures, which detect automatically the best type of CPD for each node. These modifications illustrate how a score and search algorithm, or a constraint-based learning algorithm can be adapted to learn SPBNs; (3) creation of a new learning operator for the score and search learning algorithms; (4) definition of a learning score inspired by cross-validation, which is score decomposable; (5) availability of the source code of the SPBN framework; (6) a real use case of bearing degradation monitoring.

The paper is organized as follows. Section~\ref{sec:preliminaries} introduces the concepts of continuous Bayesian networks and parametric/nonparametric models by reviewing several previous works. In Section~\ref{sec:spbn}, the proposed approach is explained in detail, including the definition of a model with several useful theoretical propositions and the adaptation of two learning algorithms. Section~\ref{sec:experiments} provides the discussion on the experimental results obtained by testing on artificial datasets, on datasets sampled from Gaussian Bayesian networks, on datasets extracted from the UCI repository, and on bearing degradation datasets. Section~\ref{sec:conclusion} concludes the paper and outlines future research directions.

\section{Background}\label{sec:preliminaries}

Hereinafter, we use the following notation: capital letters are used to denote random variables, $X$, with boldface letters representing the vectors of random variables, $\mathbf{X} = (X_{1}, \ldots, X_{n})$. A subscript applied to a vector is used to index it with a single index, $X_{i}$, or a set of indices, $\mathbf{X}_{S}$, with $S \subseteq \{1, \ldots, n\}$. Lowercase letters are utilized to denote the values of random variables, $x_{i}$, and of random variable vectors, $\mathbf{x}_{S}$, with $S \subseteq \{1, \ldots, n\}$.

A Bayesian network is a tuple $\mathcal{B} = (\mathcal{G}, \thetab)$, where $\mathcal{G} = (V, A)$ is a directed acyclic graph (DAG) with a set of nodes $V = \{1,\ldots, n\}$ and a set of directed arcs $A \subseteq V \times V$. A Bayesian network represents a factorization of a joint probability distribution $P(\mathbf{x})$ over a vector of random variables $\mathbf{X} = (X_{1}, \ldots, X_{n})$. The set of nodes $V$ index the vector of random variables so that $\mathbf{X}_{V} = \mathbf{X}$. The probability distribution $P$ factorizes over a graph $\mathcal{G}$ as:
\begin{equation}
P(x_{1}, \ldots, x_{n}) = \prod_{i=1}^{n} P\left(x_{i} \mid \mathbf{x}_{\pa(i)}\right),
\label{eq:bayesian_factorization}
\end{equation}
where $\pa(i)$ is the set of parents of the $i$ node in graph $\mathcal{G}$, and $P\left(x_{i} \mid \mathbf{x}_{\pa(i)}\right)$ is the CPD of $X_{i}$ variable given $\mathbf{x}_{\pa(i)}$. The set of CPDs $\thetab = \{P\left(x_{i} \mid \mathbf{x}_{\pa(i)}\right),$ $ i = 1,\ldots, n\}$ includes the CPD of all variables.

A Bayesian network factorizes a distribution $P$ according to Equation~(\ref{eq:bayesian_factorization}), which usually allows representing a distribution with fewer parameters. Moreover, conditional independencies between random variables can be read directly from graph $\mathcal{G}$ using the d-separation criterion \cite{Koller}.

\subsection{Parametric Continuous Bayesian Networks}

A continuous Bayesian network is a Bayesian network in which a CPD of each variable is represented using a continuous probability distribution. A Gaussian Bayesian network (GBN) \cite{Shachter1989, Geiger1994} is a Bayesian network in which all CPDs are defined using a linear Gaussian CPD. A linear Gaussian CPD assumes a conditional normal distribution for variable $X_{i}$ given $\mathbf{X}_{\pa(i)}$, as well as a linear relationship between $X_{i}$ and $\mathbf{X}_{\pa(i)}$. It can be considered as an alternative representation of a multivariate Gaussian distribution. 
As particular distributions do not fit well to a multivariate Gaussian distribution (for example, bimodal distributions), GBNs perform poorly when modeling such distributions. Furthermore, as linear Gaussian CPDs are linear in nature, they are inapplicable to represent nonlinear interactions between random variables.

Concerning the assumption of data normality in GBNs, several authors have attempted to relax it. A mixture of Bayesian networks (MoBN) \cite{Thiesson97learningmixtures} can be used to represent mixtures of multivariate Gaussian distributions. Alternatively, mixtures of truncated exponentials (MTE) \cite{MixturesTruncatedExponentials} rely on piecewise-defined exponential functions to model the CPDs of a Bayesian network. As an extension of this idea, the mixtures of polynomials (MoP) \cite{MixturesPolynomials} and truncated basis functions (MoTBF) \cite{MixturesTruncatedBasisFunctions} have been introduced to define CPDs using the piecewise functions with the polynomial and sum-of-basis function components, respectively.

\subsection{Nonparametric Continuous Bayesian Networks}\label{sec:background_npbn}

An alternative approach to avoid assuming data normality is to combine Bayesian network models with other nonparametric estimation models that are more flexible compared with parametric estimation models as they do not assume any type of parametric distribution. Instead, the complexity of the nonparametric representation may grow according to the sample size. As no parametric distribution is assumed, the density estimation error can be decreased when more data are available. This contradicts a parametric model, because with an incorrect parametric assumption the inherent bias cannot be reduced increasing the amount of data \cite{Scott2015Bandwidth}. Note that nonparametric estimation models provide worse convergence rates of the squared error of the density estimation (see below) compared to parametric estimation models, using the same sample size. For this reason, parametric estimation models are deemed more appropriate if data meet parametric assumptions.

In \cite{KDEBayesNetworkFirst}, the estimation of a conditional density of each continuous variable is defined as a ratio of the two estimations of joint densities: $\hat{f}(x_{i} \mid \mathbf{x}_{\pa(i)}) = \hat{f}(x_{i}, \mathbf{x}_{\pa(i)}) / \hat{f}(\mathbf{x}_{\pa(i)})$, where joint distributions are estimated using the kernel density estimation (KDE) \cite{Scott2015Bandwidth, Silverman1986KDE, Wand1994KDE} with multivariate Gaussian kernels. We denote this type of Bayesian networks as KDEBNs. KDE is a nonparametric estimation model defined as:
\begin{equation}
	\hat{f}(\mathbf{x}) = \frac{1}{N}\lvert\mathbf{H}\rvert^{-1/2}\sum\limits_{j=1}^{N} K\left(\mathbf{H}^{-1/2}(\mathbf{x} - \mathbf{x}^{j})\right) = \frac{1}{N}\sum\limits_{j=1}^{N} K_{\mathbf{H}}\left(\mathbf{x} - \mathbf{x}^{j}\right),\label{eq:KDE_model}
\end{equation}
where $\mathbf{x}^{j}$ is the $j$-th instance among $N$ instances in the training set and, $\mathbf{H}$ is a symmetric positive definite $n \times n$ matrix called bandwidth. A bandwidth matrix can be used to define the smoothness of density estimation. Higher values in a bandwidth produce smoother densities, while smaller values generate wiggly density estimations. $K(\mathbf{x})$ is an $n$-variate kernel function that integrates to 1, and $K_{\mathbf{H}}(\mathbf{x}) = \lvert\mathbf{H}\rvert^{-1/2}K\left(\mathbf{H}^{-1/2}\mathbf{x}\right)$. A Gaussian kernel, $K(\mathbf{x}) = \frac{1}{(2\pi)^{n/2}}\exp\left(-\frac{1}{2}\mathbf{x}^{T}\mathbf{x}\right)$, is typically used as it is a well-known distribution with notable theoretical properties. Namely, when a Gaussian kernel is used, the KDE model is equivalent to a Gaussian mixture model with an equiprobable component for each training instance.

The KDE model has also been applied to Bayesian network classifiers. Accordingly, a naive Bayes classifier is learned in \cite{John1995} using KDE to deal with continuous predictors. In \cite{Perez2009}, more flexible Bayesian network classifiers are learned defining CPDs as the ratio of the joint densities of KDE models.

The mean squared error, $\text{MSE}(\hat{f}(\mathbf{x})) = \mathbb{E}\left[(f(\mathbf{x}) - \hat{f}(\mathbf{x}))^{2}\right]$, of the KDE model is $O\left(N^{-4/(4+n)}\right)$, which is a worse convergence rate compared with the usual parametric convergence rate of $O(N^{-1})$. The error convergence rate of the KDE model also indicates that the performance of the KDE model deteriorates with an increase in dimensionality. Due to this reason, the use of KDE models in a Bayesian network is reasonable because they factorize the joint distribution. Therefore, it is better to estimate the joint distribution using multiple KDE models with a few variables in the CPDs of the Bayesian network, taking advantage of the factorization, than a single KDE model containing all the variables. Furthermore, we note that KDE is a point-wise consistent estimator\footnote{A density estimator is point-wise consistent if $\hat{f}(\mathbf{x}) \rightarrow f(\mathbf{x})$ for all $\mathbf{x}$ when $N \rightarrow \infty$.} and a consistent estimator in the mean squared error\footnote{A density estimator is a consistent estimator in the mean squared error if $\text{MSE}(\hat{f}(\mathbf{x})) \rightarrow 0$ when $N \rightarrow \infty$.}, under some mild conditions~\cite{Silverman1986KDE}.

Other nonparametric models have been combined with Bayesian networks. Friedman and Nachman \cite{GaussianProcess} have employed a Gaussian process priors to learn a functional relationship between variables. In \cite{NonparametricBN}, a Bayesian network has been trained using a nonparametric mixture model\footnote{Nonparametric mixture models define mixtures with a potentially infinite number of components.} to avoid the assumption of data normality.

\section{Semiparametric Bayesian Networks}\label{sec:spbn}

In this section, we introduce SPBNs that combine the characteristics of parametric and nonparametric Bayesian networks. First, the representation of SPBN is detailed in Section~\ref{sec:representation}. Then, two algorithms are proposed to automatically learn the structure of an SPBN from data in Section~\ref{sec:learning}. Finally, the asymptotic time complexity of all learning procedures is analyzed in Section~\ref{sec:time_complexity}.

\subsection{Representation of Semiparametric Bayesian Networks}\label{sec:representation}

SPBNs are composed of parametric and nonparametric CPDs. In this section, we describe their representation.

For the parametric CPDs, we used the well-known linear Gaussian (LG) CPDs used in GBNs, as they are easy to train and usually offer good performance when there is a linear relationship between variables. The nonparametric CPDs are represented as the ratio of two joint KDE models, as in the KDEBNs. We denote this type of CPDs as conditional kernel density estimation (CKDE) distributions.

\subsubsection{Linear Gaussian}

Let $X_{i}$ be a random variable with an LG conditional distribution; then it is assumed that:
\begin{equation}
X_{i} = \beta_{i0} + \sum_{k \in \pa(i)} \beta_{ik}X_{k} + \epsilon_{i},\quad \text{with } \epsilon_{i} \sim \mathcal{N}\left(0, \sigma_{i}^{2}\right),\label{eq:linear_random_variable}
\end{equation}
where $\mathcal{N}(\mu; \sigma^{2})$ is the normal probability density function with mean $\mu$ and variance $\sigma^{2}$; $\beta_{ik}$ is the regression coefficient for variable $X_{k}$, $k\in\pa(i)$, in the linear regression of variable $X_{i}$, and $\beta_{i0}$ is its intercept; $\epsilon_{i}$ is an error random variable with variance $\sigma_{i}^{2}$.

That is, random variable $X_{i}$ is a linear combination of its parent random variables plus a normal error. We assume that error variable $\epsilon_{i}$ is conditionally independent of all regressor variables $X_{k}$, for $k\in\pa(i)$. From Equation~(\ref{eq:linear_random_variable}), the conditional distribution of $X_{i}$ given its parents is the Gaussian distribution.

\begin{definition}
LG CPD. Let $X_{i}$ be a random variable following an LG conditional distribution; then, the conditional distribution of $X_{i}$ given $\mathbf{X}_{\pa(i)}$ can be formulated as:
\begin{equation}
f(x_{i}\mid \mathbf{x}_{\pa(i)}) = \mathcal{N}\left(\beta_{i0} + \sum_{k \in \pa(i)} \beta_{ik}x_{k}, \sigma_{i}^{2}\right).\label{eq:conditional_random_variable}
\end{equation}
\label{def:LG}
\end{definition}

In GBNs, random variable $X_{i}$ follows an unconditional normal distribution, similarly as each parent random variable. This can be easily derived from Equation~(\ref{eq:linear_random_variable}), because the linear combination of normal random variables is also normally distributed. Moreover, the unconditional distribution of multivariate random variables ($X_{i}$, $\mathbf{X}_{\pa(i)}$) and $\mathbf{X}$ is also multivariate normal distribution.

It should be noted that SPBNs do not make assumptions about the normality of parent random variables (see below). Therefore, in an SPBN, the unconditional distribution of random variables following the LG conditional distribution $X_{i}$ and ($X_{i}$, $\mathbf{X}_{\pa(i)}$) may not be necessarily normal.

However, if the assumption of the normality of parents holds, then the unconditional distribution of $X_{i}$ is exactly the same as in GBNs with the same unconditional distribution for parent random variables and the same parameter values.

\begin{proposition}
\label{prop:gaussian_networks}
An SPBN in which all the nodes are LG CPDs is equivalent to a GBN with the same arcs and parameter values.
\begin{proof}
The proof is straightforward, as, by definition, a GBN is a Bayesian network in which all CPDs are LG CPDs.\qed
\end{proof}
\end{proposition}

Based on Proposition~\ref{prop:gaussian_networks}, we can easily deduce that every possible GBN is contained in the class of SPBNs.

\subsubsection{Conditional Kernel Density Estimation}

\begin{definition} CKDE CPD. Let $X_{i}$ be a random variable following a CKDE conditional distribution; then, the conditional distribution of $X_{i}$ given $\mathbf{X}_{\pa(i)}$ is defined as:
\begin{equation}
\hat{f}(x_{i}\mid \mathbf{x}_{\pa(i)}) = \frac{\hat{f}(x_{i}, \mathbf{x}_{\pa(i)})}{\hat{f}(\mathbf{x}_{\pa(i)})},
\label{eq:CKDE_bayes}
\end{equation}
where $\hat{f}(x_{i}, \mathbf{x}_{\pa(i)})$ and $\hat{f}(\mathbf{x}_{\pa(i)})$ are KDE models, as defined in Equation~(\ref{eq:KDE_model}).
\label{def:CKDE}
\end{definition}

This CPD does not assume any underlying distribution in data by modeling multivariate random variable ($X_{i}$, $\mathbf{X}_{\pa(i)}$) using a nonparametric model. In this study, we use a Gaussian kernel for each CKDE; however, any kernel can be applied if a valid bandwidth matrix can be estimated (see Section~\ref{sec:learning_parameter}).

Therefore, the following proposition holds:

\begin{proposition}
\label{prop:kde_networks}
An SPBN in which all variables follow a CKDE CPD is equivalent to a KDEBN model with the same arcs and bandwidth matrices and is trained on the same data.

\begin{proof}
The proof is straightforward, as, by definition, a KDEBN is a Bayesian network in which all CPDs are CKDE CPDs.\qed
\end{proof}
\end{proposition}

Based on Proposition~\ref{prop:kde_networks}, we can easily deduce that every possible KDEBN is contained in the class of SPBNs.

\subsubsection{Graph Structure}

\begin{figure}
\begin{center}
\includegraphics[width=0.18473\linewidth]{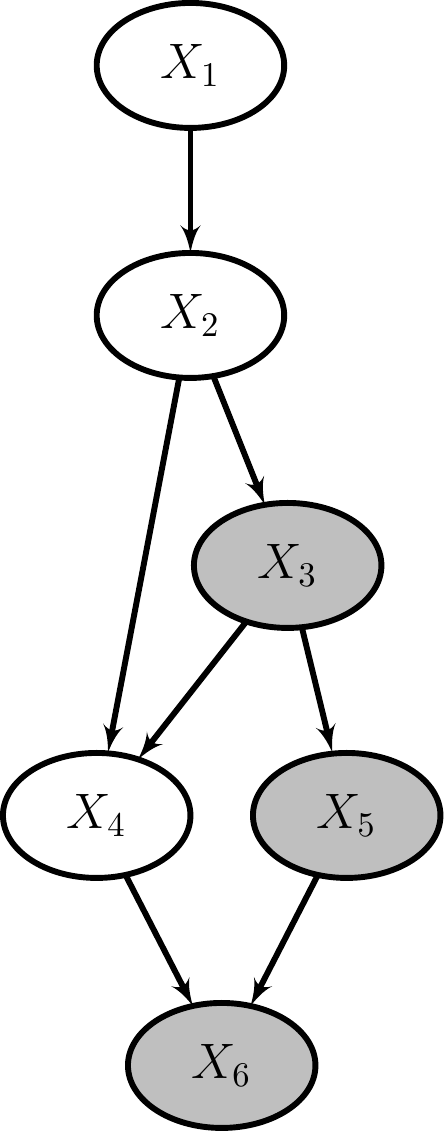}
\end{center}
\caption{Structure of an example of SPBN. White nodes are of the LG type, and gray shaded nodes are of the CKDE type.}\label{fig:example_spbn}
\end{figure}

In the SPBN model, the graph contains the type of each node, which determines the type of the corresponding CPD. There are no restrictions on the arcs, so the parent sets of each variable can be of different types: only LG parents, only CKDE parents or a mix of both options. Figure~\ref{fig:example_spbn} illustrates an example of SPBN. Here, the LG and CKDE nodes are depicted using white and gray shaded nodes, respectively. As can be seen, there are different combinations of parent node types. Analyzing this structure, we can guarantee that the unconditional probability distribution of random variables $X_{1}$, $X_{2}$ and ($X_{1}$, $X_{2}$) is Gaussian. However, the unconditional probability distribution of remaining random variables cannot be known from the structure. The conditional distribution of variables $X_{1}$, $X_{2}$ and $X_{4}$ is known to be Gaussian, and their relationship with their parents is linear. The interpretability of the structure can serve as a useful tool to extract knowledge from an SPBN learned from data.

\subsection{Learning of Semiparametric Bayesian Networks}\label{sec:learning}

In this section we present a procedure to learn the structure and parameters of a SPBN from data. There are two main types of methods to learn the structure of a Bayesian network from data. The constraint-based approaches are based on performing conditional independence tests and reconstructing a Bayesian network structure by representing the same tested conditional independences as accurately as possible~\cite{PCAlgorithm}. The score and search approaches rely on defining a scoring function that measures how well the Bayesian network structure fits to the training data. Then, the structure learning problem turns into the search for the Bayesian network structure that scores the best.

In this work, we adapt a score and search procedure (HC), and also a constraint-based procedure (PC) to illustrate how both types of methods can be used to learn an SPBN.

\subsubsection{Parameter Learning}\label{sec:learning_parameter}

Let us assume that the structure of a SPBN is fixed. That is, the set of arcs of a graph and the type of CPD of each node are known. Then, the parameters of each node CPD need to be estimated to complete the model. We will use standard techniques in the literature to learn the parameters, taking advantage of the locality of each CPD. A common approach for parameter estimation is to employ the maximum likelihood criterion.

This criterion is used to select the parameters maximizing the likelihood function. Let $\mathcal{D} = \{\mathbf{x}^{1}, \ldots, \mathbf{x}^{N}\}$, with $\mathbf{x}^{j} = (x_{1}^{j},\ldots,x_{n}^{j})$ be a set of $N$ independent and identically distributed training instances, and $\thetab$ denote a particular set of parameters. Then, the likelihood function is defined as the density assigned to the training data $\mathcal{D}$ by the Bayesian network:
\begin{equation}
f(\mathcal{D} \mid \thetab, \mathcal{G}) = \prod_{j=1}^{N} f(\mathbf{x}^{j}\mid \thetab, \mathcal{G}) = \prod_{j=1}^{N}\prod_{i=1}^{n} f(x^{j}_{i}\mid \thetab_{i}, \mathbf{x}^{j}_{\pa(i)}),
\label{eq:likelihood}
\end{equation}
where $\thetab_{i}$ is the set of parameters for a CPD of node $i$. The second equality holds, as the set of parameters $\thetab_{i}$ and $\thetab_{k}$ ($i\neq k$) are disjoint; namely, they do not share any parameter. This property is called global likelihood decomposition and indicates that maximizing $\prod\limits_{j=1}^{N}f(x^{j}_{i}\mid \thetab_{i}, \mathbf{x}^{j}_{\pa(i)})$ independently for each $i$ maximizes $f(\mathcal{D} \mid \thetab, \mathcal{G})$. Generally, the log of the likelihood, $\mathcal{L}(\mathcal{G}, \thetab : \mathcal{D})$, called also log-likelihood, is optimized, as it provides better numerical precision and can be expressed as a sum of terms:
\begin{equation}
\mathcal{L}(\mathcal{G}, \thetab: \mathcal{D}) = \sum_{j=1}^{N}\sum_{i=1}^{n} \log f(x^{j}_{i}\mid \thetab_{i}, \mathbf{x}^{j}_{\pa(i)}).
\label{eq:loglikelihood}
\end{equation}

For random variable $X_{i}$ following an LG conditional distribution, the set of parameters is defined as: $\thetab_{i} = \left\lbrace\beta_{i0}, (\beta_{ik})_{k \in \pa(i)}, \sigma_{i}^{2}\right\rbrace$. Under the assumption of a linear relationship defined in Equation~(\ref{eq:linear_random_variable}), it can be demonstrated that the maximum likelihood estimations can be obtained using an ordinary least squares estimator \cite{linear_regression_fox}. For this reason, we employ the ordinary least squares estimate for $\thetab_{i}$ if $X_{i}$ follows an LG conditional distribution.

A CKDE conditional distribution is composed of two nonparametric distributions: $\hat{f}(x_{i}, \mathbf{x}_{\pa(i)})$ and $\hat{f}(\mathbf{x}_{\pa(i)})$. For each nonparametric model, two bandwidth matrices $\mathbf{H}_{i}$ (for $\hat{f}(x_{i}, \mathbf{x}_{\pa(i)})$) and $\mathbf{H}_{i}^{-}$ (for $\hat{f}(\mathbf{x}_{\pa(i)})$) need to be estimated. However, both bandwidth matrices cannot be computed independently, as the conditional distribution $\hat{f}(x_{i}\mid \mathbf{x}_{\pa(i)})$ must integrate to 1. Then, a valid selection of $\mathbf{H}_{i}$ and $\mathbf{H}_{i}^{-}$ ensures that:
\begin{equation*}
\hat{f}(\mathbf{x}_{\pa(i)}) = \int_{-\infty}^{\infty}\hat{f}(x_{i}, \mathbf{x}_{\pa(i)})dx_{i},\ \forall x_{i}, \mathbf{x}_{\pa(i)}.
\end{equation*}

Expanding the expression (without constant terms $\frac{1}{N}$) and assuming a Gaussian kernel in both KDE models, we can formulate the following statement:
\begin{equation*}
\sum\limits_{j=1}^{N} \mathcal{N}\left(\mathbf{x}_{\pa(i)} - \mathbf{x}_{\pa(i)}^{j}, \mathbf{H}_{i}^{-}
\right) = \int_{-\infty}^{\infty}\sum\limits_{j=1}^{N} \mathcal{N}\left(\begin{bmatrix}x_{i}\\
					 \mathbf{x}_{\pa(i)}
	 \end{bmatrix} - \begin{bmatrix}x_{i}^{j}\\
					 \mathbf{x}_{\pa(i)}^{j}
					 \end{bmatrix}, \mathbf{H}_{i}
\right)dx_{i}.
\end{equation*}

Using Fubini's theorem, we have that:
\begin{equation*}
\sum\limits_{j=1}^{N} \mathcal{N}\left(\mathbf{x}_{\pa(i)} - \mathbf{x}_{\pa(i)}^{j}, \mathbf{H}_{i}^{-}
\right) = \sum\limits_{j=1}^{N} \int_{-\infty}^{\infty} \mathcal{N}\left(\begin{bmatrix}x_{i}\\
					 \mathbf{x}_{\pa(i)}
	 \end{bmatrix} - \begin{bmatrix}x_{i}^{j}\\
					 \mathbf{x}_{\pa(i)}^{j}
					 \end{bmatrix}, \mathbf{H}_{i}
\right)dx_{i}.
\end{equation*}

The marginalization of a multivariate normal distribution is also a multivariate normal distribution with the means and covariance matrix of non-marginalized variables. Let $\mathbf{H}_{i}$ be a block matrix:
\begin{equation*}
\mathbf{H}_{i} = \begin{bmatrix} 
				\alpha_{i} & \boldsymbol{\nu}_{i}^{T}\\
				\boldsymbol{\nu}_{i} & \mathbf{M}_{i}
			  \end{bmatrix}.
\end{equation*}

Then, a CKDE CPD with Gaussian kernels only integrates to 1 if and only if $\mathbf{H}_{i}^{-} = \mathbf{M}_{i}$. Accordingly, a CKDE CPD can be fitted estimating only bandwidth matrix $\mathbf{H}_{i}$.

Here, $\mathbf{H}_{i}$ cannot be estimated using the maximum likelihood because the training data constitute a part of the KDE model. From Equation~(\ref{eq:KDE_model}), when calculating the likelihood for instance $\mathbf{x}^{i}$, there is a term $K_{\mathbf{H}}\left(\mathbf{0}\right)$. This is the largest term in the sum of Equation~(\ref{eq:KDE_model}) and is maximized for bandwidth $\mathbf{H}_{i}$ with determinant $\lvert\mathbf{H}_{i}\rvert \rightarrow 0$, leading to a likelihood approaching infinity. Generally, KDE models are trained using other error criteria, such as, for example, minimizing the mean squared error. In this study, we employ the normal reference rule~\cite{Scott2015Bandwidth}, a fast rule-of-thumb estimator based on minimizing the mean integrated squared error. The normal reference rule scales a sample covariance matrix according to a factor that depends on the number of instances and dimensionality:
\begin{equation}
	\mathbf{H}_{i} = N^{-2 / (\lvert\pa(i)\rvert+5)}\hat{\boldsymbol{\Sigma}},
\end{equation}
where $\hat{\boldsymbol{\Sigma}}$ is the sample covariance matrix of random variables $X_{i}$ and $\mathbf{X}_{\pa(i)}$. In this study, we use a full covariance matrix to estimate $\mathbf{H}_{i}$. Other previous works \cite{KDEBayesNetworkFirst, Perez2009} used diagonal matrices (also called kernel \citep{Scott2015Bandwidth}), i.e. $\mathbf{H} = h\cdot\text{diag}(s_{1},\ldots,s_{n})$, where $\text{diag}()$ defines a diagonal matrix, $h$ is a smoothing parameter, and $s_{i}$ is the standard deviation of $X_{i}$. Thus, we consider our model more flexible than previous approaches. Testing other bandwidth selection methods is left as a future research direction.

\subsubsection{Greedy Hill-Climbing}\label{sec:learning_structure}

The HC algorithm is aimed at optimizing the structure of a network by moving through the space of structures applying operators that, generally, make small and local changes on a candidate structure. The set of operators defines a neighborhood set of candidate structures. At each step, the operator that produces the largest improvement in score is applied to generate a new candidate structure. The algorithm runs until a local optimum (which can be a global optimum) is reached.

Generally, three operators are utilized in HC: arc addition, arc removal, and arc reversal. In SPBNs, the structure is composed of arcs in a graph and the types of nodes, namely, LG or CKDE conditional distributions. Then, a new operator is added into the HC algorithm to learn SPBNs: node type change. This operator is denoted by \textsc{Type-Change}($i$), where $i$ is a node index. The \textsc{Type-Change} operator can change the type of a single node in a graph. That is, an LG node can be changed to be a CKDE node, and vice versa.

The definition of a score function is an important part of the score and search algorithms. The maximum of the log-likelihood function defined in Equation~(\ref{eq:loglikelihood}), namely, $\max \mathcal{L}(\mathcal{G}, \thetab : \mathcal{D})$, is a score function. However, it can be shown that maximum log-likelihood leads to overfitting, by adding too many arcs in the structure. Thus, the Bayesian Information Criterion (BIC) includes a penalization term to the log-likelihood score that depends on the number of parameters~\cite{Koller}. Under particular assumptions, including $N\rightarrow \infty$, the BIC score can be considered as an approximation to a Bayesian score. A Bayesian score allows incorporating prior probability distributions into parameters $\thetab$ and structure $\mathcal{G}$. For GBNs, a commonly used Bayesian score is the BGe score~\cite{Geiger1994}. 

However, for an SPBN, any score including the log-likelihood of the training data, such as the maximum likelihood score or BIC, are inappropriate because the training data constitute part of the KDE model. As shown in Section~\ref{sec:learning_parameter}, for each training instance, there would be a term $K_{\mathbf{H}}(\mathbf{0})$ defined in Equation~(\ref{eq:KDE_model}). Considering that the maximum argument of the kernel function is always $\mathbf{0}$, $\argmax_{\mathbf{x}} K_{\mathbf{H}}(\mathbf{x}) = \mathbf{0}$, the log-likelihood of the training data overestimates the goodness of a model on unseen data. This is explained because the probability that the unseen data is exactly the same as the training data is 0, so there will not be $K_{\mathbf{H}}(\mathbf{0})$ terms almost surely while evaluating the log-likelihood of the unseen data. Furthermore, the number of free parameters, $\text{Dim}(\mathcal{G})$, cannot be measured, because the size of the KDE model depends on the size of a training dataset rather than being specified a priori as in a parametric model. Instead, we propose to use the $k$-fold cross-validated log-likelihood:

\begin{equation}
\mathcal{S}_{\text{CV}}^{k}(\mathcal{D}, \mathcal{G}) = \sum_{m=1}^{k} \mathcal{L}(\mathcal{G}, \thetab_{\mathcal{I}_{\text{train}}^{m}} : \mathcal{D}_{\mathcal{I}_{\text{test}}^{m}}),
\label{eq:cv_loglikelihood}
\end{equation}
where $\mathcal{L}(\mathcal{G}, \thetab_{\mathcal{I}_{\text{train}}^{m}} : \mathcal{D}_{\mathcal{I}_{\text{test}}^{m}})$ is the log-likelihood of the $m$-th test fold data element in a model composed of graph $\mathcal{G}$ and parameters $\thetab_{\mathcal{I}_{\text{train}}^{m}}$. The $\thetab_{\mathcal{I}_{\text{train}}^{m}}$ parameters are estimated based on the data $\mathcal{D}_{\mathcal{I}_{\text{train}}^{m}}$ using the parameter learning procedure described in Section~\ref{sec:learning_parameter}. It is common in $k$-fold cross-validation that $\mathcal{I} = \{\mathcal{I}_{\text{test}}^{i}\}_{i=1}^{k}$ corresponds to $k$ disjoint sets of indices, and $\mathcal{I}_{\text{train}}^{i} = \bigcup_{j \neq i} \mathcal{I}_{\text{test}}^{j}$, for all $i=1,\ldots, k$. We note that the likelihood function is computed by Equation~(\ref{eq:loglikelihood}), where the contribution of each node depends on its type, as in Definition~\ref{def:LG} for LG nodes and in Definition~\ref{def:CKDE} for CKDE nodes.

Some scores have the property of decomposability \cite{Koller}. A score $\mathcal{S}$ is decomposable if it can be expressed as the sum of local score terms related to each node.

The decomposability of a score is important, as a local change in a structure only modifies a limited number of local score terms. Then, during the structure search process, the difference in scores (also called delta scores) provided by each operator can be efficiently cached. It has been demonstrated that log-likelihood, BIC, and BGe scores are decomposable~\cite{Koller,Geiger1994}. The cross-validated score in Equation~(\ref{eq:cv_loglikelihood}) is also decomposable given a selection of disjoint sets of indices $\mathcal{I}$, as it is just summing $k$ log-likelihood scores, which are themselves decomposable.

Note that the score of Equation~(\ref{eq:cv_loglikelihood}) would return different results for differents sets of indices $\mathcal{I}$ and $\mathcal{I}'$, as the training and test data used to estimate the score of each fold is different. For this reason, we need to fix a specific set of indices $\mathcal{I}$ during the learning process to make the score decomposable. However, fixing the set of indices for the cross-validated log-likelihood can induce searching for solutions that are only optimal for the specific set of indices $\mathcal{I}$. This is a type of overfitting that can be solved using the early-stopping criterion~\cite{EarlyStopping} that randomly splits the data $\mathcal{D}$ into two disjoint datasets called the training and validation sets, $\mathcal{D} = \mathcal{D}^{\text{train}} \cup \mathcal{D}^{\text{val}}$. Now, the learning process will be guided by the subset $\mathcal{D}^{\text{train}}$ and a fixed set of indices $\mathcal{I}$ over the data $\mathcal{D}^{\text{train}}$, while the subset $\mathcal{D}^{\text{val}}$ controls the overfitting to the set of indices $\mathcal{I}$. Thus, the selection of new operators in HC is performed using the score $\mathcal{S}_{\text{CV}}^{k}(\mathcal{D}^{\text{train}}, \mathcal{G})$, while the overfitting is controlled using $\mathcal{D}^{\text{val}}$ measuring the goodness of the new structure of each iteration as:

\begin{equation}
\mathcal{S}_{\text{validation}}(\mathcal{D}^{\text{train}}, \mathcal{D}^{\text{val}}, \mathcal{G}) = \mathcal{L}(\mathcal{G}, \thetab_{\mathcal{D}^{\text{train}}} : \mathcal{D}^{\text{val}}),
\end{equation}
where $\thetab_{\mathcal{D}^{\text{train}}}$ are the parameters estimated using the full training set $\mathcal{D}^{\text{train}}$. Therefore, if the new operator overfits to the fixed set of indices, $\mathcal{I}$, the log-likelihood of $\mathcal{D}^{\text{val}}$ does not improve.

The structure search can continue until the moment when a structure that improves the log-likelihood of $\mathcal{D}^{\text{val}}$ cannot be found for $\lambda$ iterations. Here, $\lambda$ is a parameter called patience. When $\lambda > 0$, the structure search is allowed proceeding to worse structures in $\mathcal{S}_{\text{validation}}$, during at most $\lambda$ iterations to explore new possible solutions. This can help to avoid local maxima, as, by the definition of local maxima, a better solution cannot be found in the neighborhood set of candidate structures, meaning that no better structures can be identified applying a single operator. To ensure greater exploration of new structures, the tabu search~\cite{TabuSearch} approach is applied while the algorithm proceeds through worse structures. Tabu search forbids applying operators that undo those recently applied, namely, adding and removing the same arc, reversing the same arc, changing the node type of the same node. The tabu search approach constrains the search space to explore different directions to escape from local maxima.

\algdef{SE}[DOWHILE]{Do}{doWhile}{\algorithmicdo}[1]{\algorithmicwhile\ #1}%
\begin{algorithm}
\caption{Greedy hill-climbing for SPBNs}
\label{alg:greedy_hill_climbing}
\begin{algorithmic}[1]
\Require Training data $\mathcal{D}$, starting structure $\mathcal{G}_{0}$, the set of operators $\mathcal{O}$, patience $\lambda$, the number of folds $k$, minimum delta $\epsilon$
\State $\mathcal{G}_{\text{best}} \leftarrow \mathcal{G}_{0}$
\State $\mathcal{G}_{\text{new}} \leftarrow \mathcal{G}_{0}$
\State $p \leftarrow 0$
\State $\text{Tabu} \leftarrow \emptyset$
\State $\mathcal{D}^{\text{train}}, \mathcal{D}^{\text{val}} \leftarrow \text{Split}(\mathcal{D})$
\Do
	\State $\mathcal{G} \leftarrow \mathcal{G}_\text{new}$
	\For {$o$ in $\mathcal{O}$}
		\If {$o$ does not reverse $o' \in \text{Tabu}$}
			\State $\mathcal{G}_{\text{candidate}} \leftarrow o(\mathcal{G})$
	  		\If {$\mathcal{S}_{\text{CV}}^{k}(\mathcal{D}^{\text{train}}, \mathcal{G}_{\text{candidate}}) > \mathcal{S}_{\text{CV}}^{k}(\mathcal{D}^{\text{train}}, \mathcal{G}_{\text{new}})$ \textbf{and}\par\hskip\algorithmicindent\hskip\algorithmicindent\phantom{i} $\mathcal{S}_{\text{CV}}^{k}(\mathcal{D}^{\text{train}}, \mathcal{G}_{\text{candidate}}) - \mathcal{S}_{\text{CV}}^{k}(\mathcal{D}^{\text{train}}, \mathcal{G}) > \epsilon$}
	  			\State $o_{\text{new}} \leftarrow o$
  			   	\State $\mathcal{G}_{\text{new}} \leftarrow \mathcal{G}_{\text{candidate}}$
			\EndIf
		\EndIf
	\EndFor
	\If {$\mathcal{S}_{\text{validation}}(\mathcal{D}^{\text{train}}, \mathcal{D}^{\text{val}}, \mathcal{G}_{\text{new}}) > \mathcal{S}_{\text{validation}}(\mathcal{D}^{\text{train}}, \mathcal{D}^{\text{val}}, \mathcal{G}_{\text{best}})$}
		\State $\mathcal{G}_{\text{best}} \leftarrow \mathcal{G}_{\text{new}}$
		\State $\text{Tabu} \leftarrow \emptyset$
		\State $p \leftarrow 0$
	\Else
		\State $\text{Tabu} \leftarrow \text{Tabu} \cup o_{\text{new}}$
		\State $p \leftarrow p + 1$
	\EndIf
	\State $\Call{Update\_Score\_Cache}{\mathcal{G}, o_{\text{new}}}$
\doWhile{$p < \lambda$}

\Return{$\mathcal{G}_{\text{best}}$}
\end{algorithmic}
\end{algorithm}

The details of implementing the HC algorithm are shown in Algorithm~\ref{alg:greedy_hill_climbing}. Lines 8-16 describe that it searches an operator maximizing the cross-validated log-likelihood of $\mathcal{D}^{\text{train}}$ and produces new structure $\mathcal{G}_{\text{new}}$. We note that the algorithm selects the operators with a delta score greater than $\epsilon$. In this work, we set $\epsilon = 0$, as it guarantees that the selection of operators results in improving $\mathcal{S}_{\text{CV}}^{k}$. If $\mathcal{G}_{\text{new}}$ improves $\mathcal{S}_{\text{validation}}$, the best structure so far has been found, and $\mathcal{G}_{\text{best}}$, is updated (lines 17-18). Thereafter, the tabu search is disabled (lines 19-20). If $\mathcal{G}_{\text{new}}$ does not improve $\mathcal{S}_{\text{validation}}$, the tabu set of forbidden operators is updated (lines 22-23). Parameter $p$ ensures that the algorithm stops when the best found structure has not further improved during $\lambda$ iterations. Exploiting the decomposability property of the cross-validated log-likelihood, the algorithm updates the cached delta scores, as described in line 25. This procedure allows modifying only the scores affected by applying operator $o_{\text{new}}$. This update function for the arc operators is well-known in the literature~\cite{Koller}, so we do not include it here. The delta score for the operator $\textsc{Type-Change}(i)$ is:

\begin{equation}
\Delta\textsc{Type-Change}(i) = \mathcal{S}_{\text{CV}}^{k}(X_{i}\mid \mathbf{X}_{\pa(i)}, \neg\text{Type}(i)) - \mathcal{S}_{\text{CV}}^{k}(X_{i}\mid \mathbf{X}_{\pa(i)}, \text{Type}(i))
\end{equation}
where $\mathcal{S}_{\text{CV}}^{k}(X_{i}\mid \mathbf{X}_{\pa(i)}, \text{Type}(i))$ is the local score of variable $X_{i}$, with parents $\mathbf{X}_{\pa(i)}$, when the type of CPD for $X_{i}$ is determined by the function $\text{Type}(i)$. The complement of function $\text{Type}(i)$ is denoted as $\neg\text{Type}(i)$, that is, $\neg\text{CKDE}$ is LG, and $\neg\text{LG}$ is CKDE. On each iteration of the HC algorithm, only a small amount of delta scores need to be updated depending on the last operator applied. If the last operator applied was an addition or removal of an arc $s \rightarrow d$, only $\Delta\textsc{Type-Change}(d)$ needs to be updated. For the reversal of the same arc, both $\Delta\textsc{Type-Change}(s)$ and $\Delta\textsc{Type-Change}(d)$ need the update. Lastly, if an operator $\textsc{Type-Change}(i)$ is applied, only $\Delta\textsc{Type-Change}(i)$ changes.

\subsubsection{PC algorithm}\label{sec:pc_algorithm}

The PC algorithm learns the structure of the Bayesian network by performing conditional independence tests to construct the graph that best captures the conditional independence relationships. The PC algorithm assumes that the underlying distribution is faithful to the Bayesian network graph, so that if two variables $X_{i}$ and $X_{j}$ are conditionally independent given a separating set of variables $\mathbf{S}_{ij}$, then the variables $X_{i}$ and $X_{j}$ must be d-separated in the graph $\mathcal{G}$ given $\mathbf{S}_{ij}$. Therefore, if no separating set $\mathbf{S}_{ij}$ can be found that makes $X_{i}$ and $X_{j}$ conditionally independent, then the nodes must be adjacent in the graph $\mathcal{G}$. The PC algorithm conducts the search for the separating sets $\mathbf{S}_{ij}$ that make all pairs of variables conditionally independent in an efficient manner. Once a skeleton is found that identifies which nodes are adjacent, the PC algorithm tries to orient the v-structures $X_{i} \rightarrow X_{k} \leftarrow X_{j}$ with $X_{i}$ and $X_{j}$ nonadjacent. A v-structure can be oriented if $X_{k} \notin \mathbf{S}_{ij}$. In this work, we used the MPC version of the PC stable algorithm~\cite{PCStable}, which is guaranteed to always return the same structure, even if the order in which the variables are presented to the algorithm changes. The MPC version performs a new search of the possible separating sets $\mathbf{S}_{ij}$ for every v-structure candidate, and only orients a v-structure if the majority of the separating sets do not contain $X_{k}$. We omit the details of the algorithm here and we refer the reader to~\cite{PCStable} for more details. The end product of the PC algorithm is a partially directed acyclic graph (PDAG) that represents the skeleton of an equivalence class. This PDAG can be converted into a DAG of that equivalence class by using a simple algorithm~\cite{pdag2dag}.

One of the key components in a constraint-based algorithm is the type of conditional independence test. A common choice is the use of the partial correlation test~\cite{PartialCorrelation1, PartialCorrelation2}, which assumes that all the variables are distributed with a multivariate Gaussian. Since the SPBN model does not assume the distribution of any variable, we also used nonparametric conditional independence tests. In particular, we tested the CMIknn~\cite{CMIknn} conditional independence test, that is based on the estimation of the mutual information with K-nearest neighbors. However, this is a permutation conditional independence test and it was too slow to finish the high quantity number of tests needed for the PC algorithm in a reasonable time. Therefore, we also tested the Randomized conditional Correlation Test (RCoT)~\cite{RCoT}, which is faster because its distribution under the null hypothesis can be approximated with less computational resources.

Finally, to learn an SPBN, it is needed to establish the best type of CPD for each variable given the DAG learned by PC. An appealing approach would be to perform a statistical normality test, such as Shapiro-Wilks, on the regression residuals of the LG CPD. However, most of the normality tests have too much power when the sample size is too large (\cite{ShapiroLimit} sets the limit at 5000 instances for Shapiro-Wilks), thus easily rejecting the null distribution of normality.

For this reason, we select the best node types with the execution of the HC algorithm described in Algorithm~\ref{alg:greedy_hill_climbing}, but allowing only the operator $\textsc{Type-Change}$. This ensures that the arc selection returned by PC is not modified by HC.

\subsection{Asymptotic Time Complexity}\label{sec:time_complexity}

In this section we analyze the asymptotic time complexity of the different learning procedures. For both HC and PC algorithm, the execution time depends on the number of iterations needed. Usually, this number of iterations cannot be known in advance, since it depends on the starting model, the global optimum, the local optima present in the search path, the possible innacuracies caused by the score function or the conditional independence tests, and many other factors. The score function and the conditional independence tests are always the most computationally demanding elements of a learning algorithm as we will show in the following analysis.

In the HC algorithm, the set of arc operators contains $n(n-1)$ different operators for each graph (although some of them may be innaplicable because of the acyclicity constraint). Moreover, there are always $n$ different \textsc{Type-Change} operators. The delta score of all these operators can be calculated with $n(n+1)$ evaluations of the score function, by caching the local score of each node in advance. Therefore, at the start of the HC algorithm, the number of score evaluations is quadratic on the number of nodes because the delta score of all operators is needed. The update of the delta scores after each iteration depends on the number of affected local scores, which can be $1$ (arc addition, arc removal and node type change) or $2$ (arc reversal).  In the former case, only $n + \lvert\pa(i)\rvert - 1$ arc operators and $1$ \textsc{Type-Change} operator change their delta score, where $X_{i}$ is the affected local score node. This update can be completed with $n$ score function evaluations (taking advantage of cached delta scores). In the latter case, $2n + \lvert\pa(i)\rvert + \lvert\pa(j)\rvert - 3$ arc operators and 2 \textsc{Type-Change} operator change their delta score, where $X_{i}$ and $X_{j}$ are the involved nodes in the arc reversal. This update can be completed with $2n$ score function evaluations. This analysis shows that, thanks to the decomposability of the score, the complexity decreases from quadratic to linear in the number of nodes for each iteration of the HC. In addition, to update the validation score, we only need $1$ (arc addition, arc removal and node type change) or $2$ (arc reversal) evaluations for each iteration.

We now present an analysis of the complexity of the score functions to compute the local score of node $X_{i}$ with parents $\mathbf{X}_{\pa(i)}$. The complexity of the cross-validated score function (Equation~\ref{eq:cv_loglikelihood}) is of the form $O(kT)$, where $T$ is the cost of parameter learning and log-likelihood evaluation for each fold. This complexity is different for the LG and CKDE CPDs. Let $L = N/k$ and $J = N - L$ be the number of test and train instances on each fold respectively. For the LG CPD, it is necessary to find a least squares estimate, which has a complexity of $O\left(J\lvert\pa(i)\rvert^{2}\right)$. Once the least squares estimate is found, the log-likelihood of the test instances can be computed with complexity $O(L\lvert\pa(i)\rvert)$. Since $J \geq L$ and usually $J \gg L$, the complexity of the least squares estimate dominates the complexity of the log-likelihood evaluation. Therefore, the complexity of the cross-validated score function for LG CPDs is $O(kJ\lvert\pa(i)\rvert^{2})$.

The CKDE CPD requires evaluating $LJ$ multivariate Gaussian probability density functions (with dimensionality $(\lvert\pa(i)\rvert + 1)$) for each fold. Each Gaussian evaluation has a complexity of $O\left(\left(\lvert\pa(i)\rvert + 1\right)^{2}\right)$ if the inverse and the determinant of the bandwidth matrix $\mathbf{H}$ are calculated in advance, which takes $O\left(\left(\lvert\pa(i)\rvert + 1\right)^{3}\right)$. Therefore, the complexity of the cross-validated score function for CKDE CPDs is $O(kLJ\lvert\pa(i)\rvert^{2})$ or $O(NJ\lvert\pa(i)\rvert^{2})$. This can be expressed with a looser bound as $O(N^{2}\lvert\pa(i)\rvert^{2})$, which suggests that the complexity is quadratic with respect to the number of instances and the number of parents. Furthermore, we know that the complexity increases with $k$, so the less demanding setting is $k=2$ and the most costly setting is $k=N-1$. In this work, we use $k=10$ on all the experiments. This complexity might be probably reduced using an approximation such as random Fourier features~\cite{RandomFourier}, but we leave that approach as future work. In practice, this is an embarranssingly parallel problem~\cite{Multiprocessing} because each multivariate Gaussian probability density function can be executed independently. PyBNesian implements this parallel problem using OpenCL~\cite{OpenCL} to enable GPU acceleration, which significantly speeds up the execution.

The complexity of the BIC score is dominated by the least squares estimation of parameter learning, so it is equal to $O(N\lvert\pa(i)\rvert^{2})$. The BGe score has a complexity of $O(N(\lvert\pa(i)\rvert + 1)^{2} + (\lvert\pa(i)\rvert + 1)^{3})$. The first term of the sum is because the sample sum of squared error of variables $\{X_{i}\} \cup \mathbf{X}_{\pa(i)}$ needs to be calculated~\cite{BGeImpl}. The second term is the cost of calculating the determinant of a square matrix of size $\lvert\pa(i)\rvert + 1$. However, we can cache the sample sum of squared errors at the start of the HC algorithm in $O(Nn^{2})$. Then, each score function evaluation has a cost of $O((\lvert\pa(i)\rvert + 1)^{3})$, which is usually preferable because, as we described before, the HC algorithm performs many score function evaluations.

The complexity of the PC algorithm is difficult to analyze because the number of iterations depends on the size of the largest separating set $\mathbf{S}_{ij}$ (assuming a perfect conditional independence test). In the first iteration, $\frac{n(n-1)}{2}$ unconditional independence tests are performed. In the subsequent iterations, the number of independence test executions depends on the conditional independences found in the previous iterations, so the number of expected conditional independence tests cannot be calculated. In addition, the number of possible separating sets $\mathbf{S}_{ij}$ between a pair of variables $X_{i}$ and $X_{j}$ is equal to ${\lvert\text{adj}_{l}(X_{i})\rvert - 1 \choose l} + {\lvert\text{adj}_{l}(X_{j})\rvert - 1 \choose l}$, where $\text{adj}_{l}(X_{i})$ is the set of adjacent nodes to node $X_{i}$ at iteration $l$. Furthermore, in the worst case, it is necessary to perform a conditional independence test for each possible separating set. Thus, the number of conditional independence test executions can grow quickly in the worst case.

We analyze now the complexity of a conditional independence test between $X_{i}$ and $X_{j}$ given a separator set $\mathbf{S}_{ij}$. The PLC independence test can be calculated using the precision matrix of the set of variables $\{X_{i}\} \cup \{X_{j}\} \cup \mathbf{S}_{ij}$. Thus, the complexity of the PLC independence test is $O(N(\lvert\mathbf{S}_{ij}\rvert + 2)^{2} + (\lvert\mathbf{S}_{ij}\rvert + 2)^{3})$, where the first term comes from the calculation of the covariance matrix and the second term from the complexity of its inversion. As in BGe, we can cache the covariance matrix information at the start of the algorithm in $O(Nn^{2})$, and then each evaluation of the conditional independence test can be performed in $O((\lvert\mathbf{S}_{ij}\rvert + 2)^{3})$. The RCoT independence test has a complexity of $O(N\lvert\mathbf{S}_{ij}\rvert)$, assuming the number of random Fourier features is fixed. This is the complexity of the computation of the random Fourier feature matrices, which is the most demanding procedure of the independence test. The authors provide a description of the complexity of RCoT~\cite{RCoT}, so we do not include more details here.

\section{Experimental Results}\label{sec:experiments}

In this section, we discuss the results of the experiments conducted on SPBNs and the comparison with alternative methods. We conducted four types of experiments depending on the input data source: synthetic data sampled by mixing linear and nonlinear functions, data sampled from GBNs, data from the UCI repository~\cite{UCIRepository}, and bearing degradation data. Finally, the execution times of all algorithms are shown in Section~\ref{sec:execution_times}.

The experiments were performed using the PyBNesian\footnote{\url{https://github.com/davenza/PyBNesian}} library. The source code of the experiments is available at \url{https://github.com/davenza/SPBN-Experiments}.

\subsection{Synthetic Data}

\begin{figure}
\begin{center}
\includegraphics[width=0.3\linewidth]{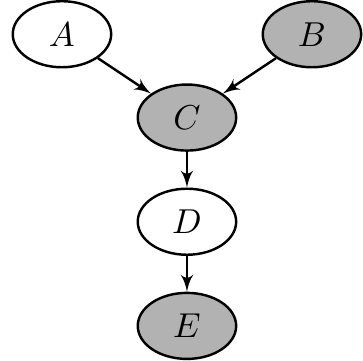}
\end{center}
\caption{Ground truth SPBN. White nodes denote the LG type, and gray shaded nodes correspond to the CKDE type.}\label{fig:true_model}
\end{figure}

In this section, we discuss the results of applying the learning algorithms introduced in Section~\ref{sec:learning} to the artificial data. We sampled the data from the following probabilistic model:

\begin{equation}
\begin{aligned}
	f(a) &= \mathcal{N}(\mu_{A} = 0, \sigma_{A}^2 = 1)\\
	f(b) &= 0.5\cdot\mathcal{N}(\mu_{B_{1}} = -2, \sigma_{B_{1}}^2 = 2) + 0.5\cdot\mathcal{N}(\mu_{B_{2}} = 2, \sigma_{B_{2}}^2 = 2)\\
	f(c | a, b) &= a \cdot b + \epsilon_{C}, \text{ where } \epsilon_{C} \sim \mathcal{N}(\mu_{\epsilon_{C}} = 0, \sigma_{\epsilon_{C}}^2 = 1)\\
	f(d | c) &= \mathcal{N}(\mu_{D} = 10 + 0.8\cdot c, \sigma_{D}^2 = 0.5)\\
	f(e | d) &= \mathcal{S}(d) + \epsilon_{E}, \text{ where } \epsilon_{E} \sim \mathcal{N}(\mu_{\epsilon_{E}} = 0, \sigma_{\epsilon_{E}}^2 = 0.5)
\end{aligned}
\label{eq:probabilistic_model}
\end{equation}
where $\mathcal{S}(x) = 1 / (1 + \exp(-x))$, is the sigmoid function. The set of conditional independences of the probabilistic model can be represented with an SPBN as in Figure~\ref{fig:true_model}. We selected this structure as it contained CKDE nodes with different types of parents. We sampled three training datasets with the different number of instances: 200, 2,000, and 10,000. In addition, we sampled another test dataset with 1,000 instances to evaluate the log-likelihood of each learned model and the ground truth model on unseen data. In such a way, we can quantitatively compare all models. To compare the learned structures, we also calculated the Hamming distance (HMD)\footnote{HMD between two graphs is used to evaluate the number of arcs that are present in a graph but not in the other graph ignoring arc directions} between the graphs corresponding to the learned models and the ground truth model. As HMD does not consider the direction of arcs, we also employed the structural Hamming distance (SHD) \cite{MaxMinHillClimbing} introduced to calculate the number of additions, removals, and reversals of arcs required to transform the DAG of the learned model into that of the ground truth model. Moreover, we computed a node type Hamming distance (THMD) measuring the number of nodes with a different node type in the learned and ground truth models. We ran the HC and PC algorithms with two different values of patience $\lambda$: 0 and 5. However, both options learned the same model, so we omit the $\lambda$ parameter in this analysis. We tested HC using a starting graph $\mathcal{G}_{0}$ without arcs and with two configurations for the type nodes: all the nodes were LG (SPBN-LG) or all the nodes were CKDE (SPBN-CKDE). We observed that in the latter case, the resulting graphs tend to be more sparse. This is reasonable because the CKDE CPDs are more flexible than LG CPDs, so it does not need as much parents to obtain a good fit to the data. For this reason, we present only the results for $\mathcal{G}_{0}$ with CKDE nodes. The PC algorithm was executed using a partial correlation test (PC-PLC) and RCoT (PC-RCoT).

We present the results in Table~\ref{tab:artificial_results}.  As expected, the ground truth demonstrated better log-likelihood compared with the learned models. Moreover, the log-likelihood and the structural accuracy improved with an increase in the number of training instances. With 10,000 instances both HC and PC-RCoT returned the ground truth structure. We can see that PC-RCoT always behaved better than PC-PLC. This is because the partial linear correlation test requires that all the variables are multivariate Gaussian, but the ground truth contains nonlinearities and some non-Gaussian distributions (such as the bimodal distribution in variable B). Furthermore, for 2,000 and 10,000 instances all the algorithms recovered the node type correctly for all the nodes.

\begin{table}
\caption{Results of training using the synthetic data of Equation~(\ref{eq:probabilistic_model}). HMD stands for the Hamming distance, SHD denotes the structural Hamming distance, and THMD corresponds to the node type Hamming distance computed between the learned model and the ground truth model.}
\label{tab:artificial_results}
\begin{center}
\begin{tabular}{llcccc}
\hline
Model & & Log-likelihood & HMD & SHD & THMD\\
\hline
Max possible value & & & 10 & 10 & 5\\
Ground truth & & $-6982.23$ & 0 & 0 & 0\\[1em]
\multirow{3}{*}{200 instances} & HC & $-7479.48$ & 0 & 0 & 2 \\
	  & PC-PLC & $-8034.62$ & 3 & 3 & 2\\
	  & PC-RCoT & $-8031.09$ & 2 & 2 & 2\\[1em]
\multirow{3}{*}{2,000 instances} & HC & $-7217.31$ & 2 & 2 & 0 \\
	  & PC-PLC & $-7827.65$ & 4 & 4 & 0\\
	  & PC-RCoT & $-7316.59$ & 1 & 1 & 0\\[1em]
\multirow{3}{*}{10,000 instances} & HC & $-7134.90$ & 0 & 0 & 0 \\
	  & PC-PLC & $-7817.06$ & 3 & 4 & 0\\
	  & PC-RCoT & $-7134.90$ & 0 & 0 & 0\\
\hline
\end{tabular}
\end{center}
\end{table}

\begin{figure}
	\begin{subfigure}[t]{0.24\linewidth}
		\centering
		\includegraphics[width=\linewidth]{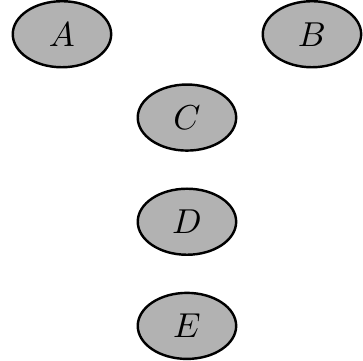}
		\caption{Start model.\\$\mathcal{S}_{\text{CV}}^{k}$: $-98286.76$\\$\mathcal{S}_{\text{validation}}$: $-24508.59$\\$\mathcal{L}(\mathcal{D}^{\text{test}})$: $-12195.51$}
	\end{subfigure}\hfil
	\begin{subfigure}[t]{0.24\linewidth}
		\centering
		\includegraphics[width=\linewidth]{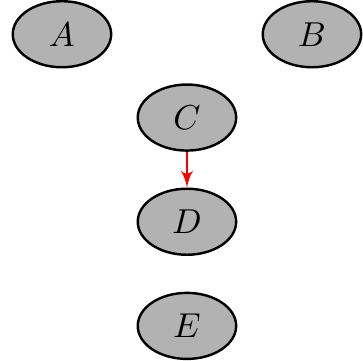}
		\caption{Iteration 1.\\$\mathcal{S}_{\text{CV}}^{k}$: $-77361.61$\\$\mathcal{S}_{\text{validation}}$: $-19307.53$\\$\mathcal{L}(\mathcal{D}^{\text{test}})$: $-9603.66$}
	\end{subfigure}\hfil
	\begin{subfigure}[t]{0.24\linewidth}
		\centering
		\includegraphics[width=\linewidth]{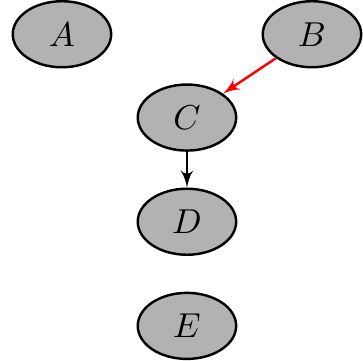}
		\caption{Iteration 2.\\$\mathcal{S}_{\text{CV}}^{k}$: $-63761.92$\\$\mathcal{S}_{\text{validation}}$: $-15864.52$\\$\mathcal{L}(\mathcal{D}^{\text{test}})$: $-7907.58$}
	\end{subfigure}\hfil
	\begin{subfigure}[t]{0.24\linewidth}
		\centering
		\includegraphics[width=\linewidth]{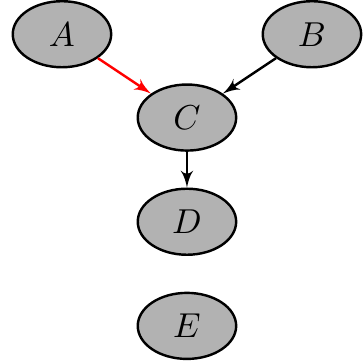}
		\caption{Iteration 3.\\$\mathcal{S}_{\text{CV}}^{k}$: $-58534.58$\\$\mathcal{S}_{\text{validation}}$: $-14577.28$\\$\mathcal{L}(\mathcal{D}^{\text{test}})$: $-7239.26$}
	\end{subfigure}\hfil
	\vspace{0.5cm}
	\begin{subfigure}[t]{0.24\linewidth}
		\centering
		\includegraphics[width=\linewidth]{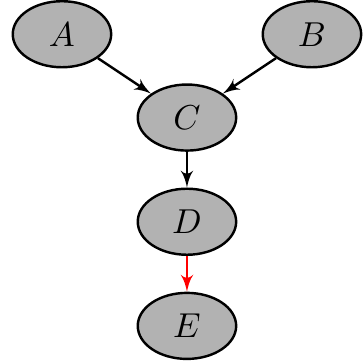}
		\caption{Iteration 4.\\$\mathcal{S}_{\text{CV}}^{k}$: $-57619.25$\\$\mathcal{S}_{\text{validation}}$: $-14342.90$\\$\mathcal{L}(\mathcal{D}^{\text{test}})$: $-7137.57$}
	\end{subfigure}\hfil
	\begin{subfigure}[t]{0.24\linewidth}
		\centering
		\includegraphics[width=\linewidth]{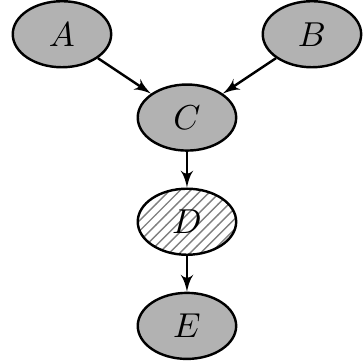}
		\caption{Iteration 5.\\$\mathcal{S}_{\text{CV}}^{k}$: $-57598.67$\\$\mathcal{S}_{\text{validation}}$: $-14323.82$\\$\mathcal{L}(\mathcal{D}^{\text{test}})$: $-7135.64$}
	\end{subfigure}\hfil
	\begin{subfigure}[t]{0.24\linewidth}
		\centering
		\includegraphics[width=\linewidth]{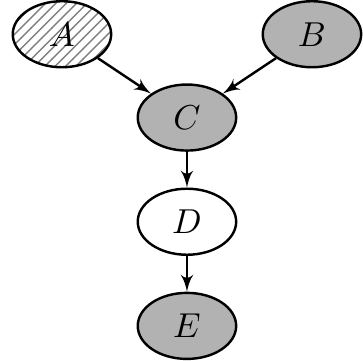}
		\caption{Iteration 6.\\$\mathcal{S}_{\text{CV}}^{k}$: $-57597.02$\\$\mathcal{S}_{\text{validation}}$: $-14320.80$\\$\mathcal{L}(\mathcal{D}^{\text{test}})$: $-7134.90$}
	\end{subfigure}\hfil
	\begin{subfigure}[t]{0.24\linewidth}
		\centering
		\includegraphics[width=\linewidth]{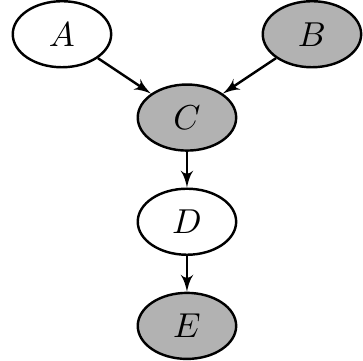}
		\caption{Final model.\\$\mathcal{S}_{\text{CV}}^{k}$: $-57597.02$\\$\mathcal{S}_{\text{validation}}$: $-14320.80$\\$\mathcal{L}(\mathcal{D}^{\text{test}})$: $-7134.90$}
	\end{subfigure}\hfil
\caption{Learning progress for the greedy hill-climbing algorithm with 10,000 training instances from the start model to the final model. An arc addition is shown with a red arc. The change from CKDE node type to LG node type is shown with striped nodes. At each iteration, the training score, $\mathcal{S}_{\text{CV}}^{k}$, the validation score, $\mathcal{S}_{\text{validation}}$, and the test log-likelihood, $\mathcal{L}(\mathcal{D}^{\text{test}})$ (we omit the $\mathcal{G},\thetab$ arguments of the $\mathcal{L}$ function), are shown.}
\label{fig:learning_progress}
\end{figure}

To illustrate the learning progress, Figure~\ref{fig:learning_progress} shows how the model changes at each iteration of the greedy hill-climbing algorithm trained with 10,000 training instances. We can see that the algorithm first added all the arcs of the structure, and then, changed node types. It is important to note that Algorithm~\ref{alg:greedy_hill_climbing} allows interleaving arc operators with node type change operators. In this specific execution, the arcs have been added first because the CKDE CPD is good enough estimating a Gaussian distribution with 10,000 instances. Thus, in the first iterations the delta scores of the arc addition operators are higher. However, in the last iterations there are no more arc changes that improve the score, so the node types are changed because they provide a refinement over the CKDE CPD.

\subsection{Data Sampled from Gaussian Networks}

Considering that GBNs constitute a special case of SPBNs, in this section, we test the SPBN learning in the case when the training data follow a multivariate Gaussian distribution. We selected four GBNs from the bnlearn~\cite{bnlearn} Bayesian network repository: ECOLI70, MAGIC-NIAB, MAGIC-IRRI, and ARTH150. We describe the properties of each Bayesian network in Table~\ref{tab:gbn_characteristics}. For each GBN, we sampled three training datasets with the different number of instances: 200, 2,000, and 10,000. Similarly as in the previous section, we sampled a test dataset of 1,000 instances to compare the log-likelihood of the trained models. The GBN models were learned using the HC algorithm with the BIC and BGe scores, and also the PC-PLC and PC-RCoT algorithms. The SPBN models were learned using the same configurations as in the previous section. In this case, the patience $\lambda = 5$ performed a little better, so we omit the results for $\lambda = 0$ here.

\begin{table}
\caption{Properties of the tested GBNs.}
\label{tab:gbn_characteristics}
\begin{center}
\begin{tabular}{lcc}
\hline
True model & Nodes & Arcs\\
\hline
ECOLI70	   & $46$   & $70$\\
MAGIC-NIAB & $44$   & $66$\\
MAGIC-IRRI & $64$   & $102$\\
ARTH150    & $107$  & $150$\\
\hline
\end{tabular}
\end{center}
\end{table}

\begin{figure}
\begin{center}
\includegraphics[width=\linewidth]{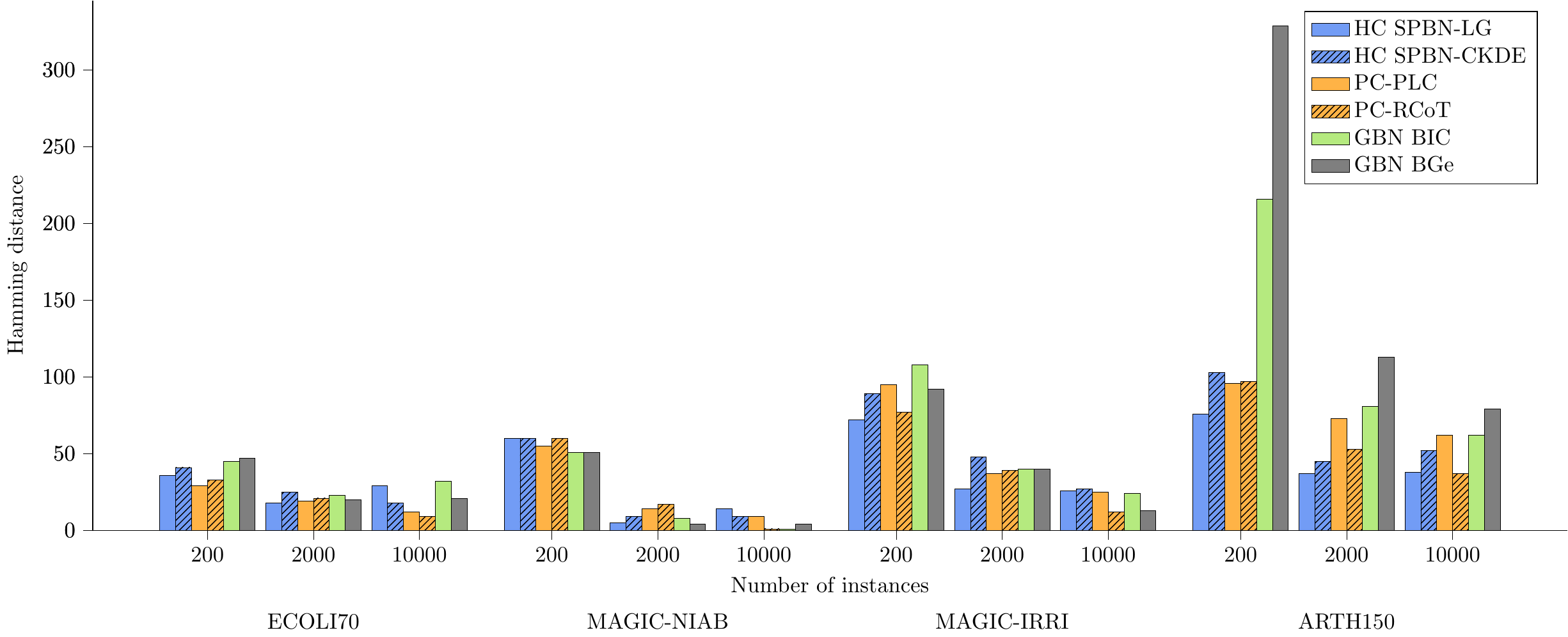}
\end{center}
\caption{HMD of the trained models to the ground truth model.}\label{fig:gbn_hmd}
\end{figure}

\begin{figure}
\begin{center}
\includegraphics[width=\linewidth]{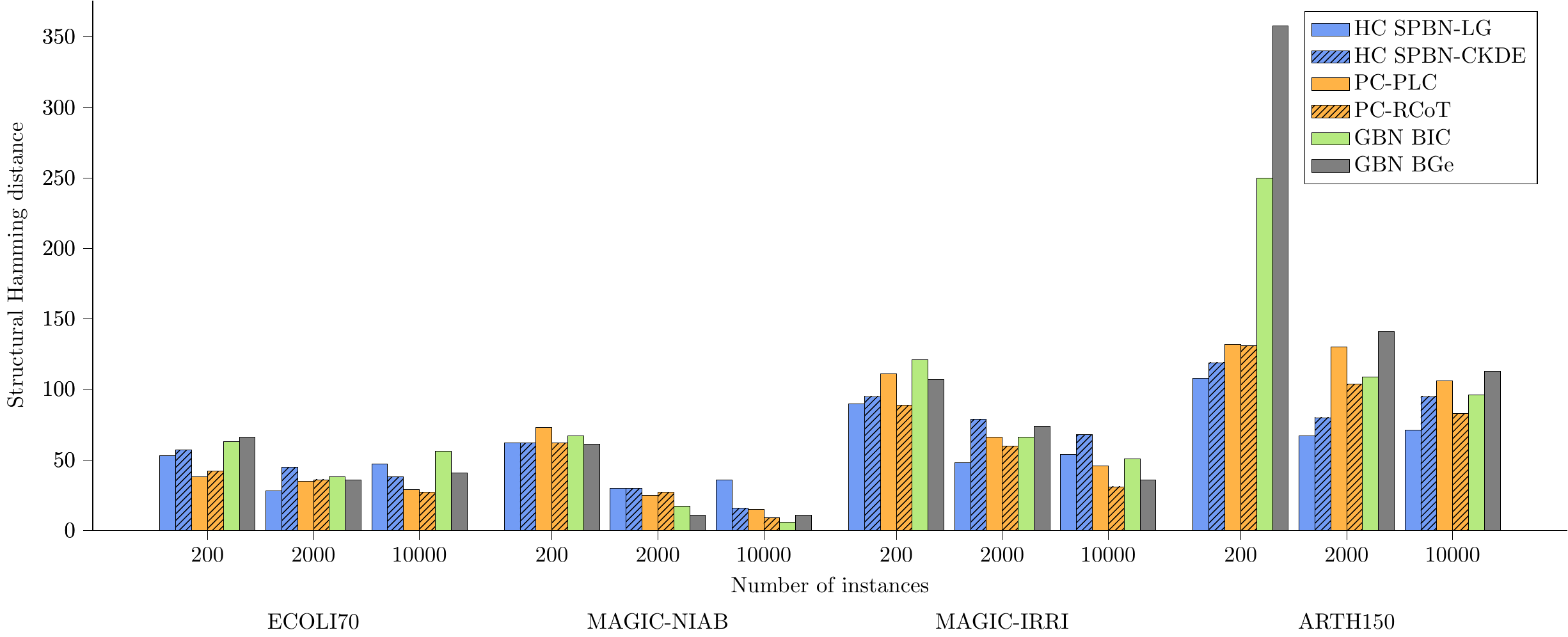}
\end{center}
\caption{SHD of the trained models to the ground truth model.}\label{fig:gbn_shd}
\end{figure}

The HMD and SHD measures of each trained model are represented in Figure~\ref{fig:gbn_hmd} and Figure~\ref{fig:gbn_shd}, respectively. All the models learned with PC have the same graph, so they are represented as PC-PLC and PC-RCoT in the figures. The SPBN-CKDE models have a worse structural accuracy than SPBN-LG. This is meaningful because in SPBN-LG the starting graph has a correct node type for all the nodes, so the algorithm only needs to optimize the arcs of the graph. Moreover, we can see that the HC algorithm for SPBNs is competitive with the PC algorithm in terms of structural accuracy. This is remarkable because the PC algorithm is known to be a better algorithm than HC in reducing the SHD~\cite{Scutari2019}.  For the PC algorithm, there are not important differences between PC-PLC and PC-RCoT. In addition, BGe score shows a specially poor structural accuracy in ARTH150.

\begin{figure}
\begin{center}
\includegraphics[width=\linewidth]{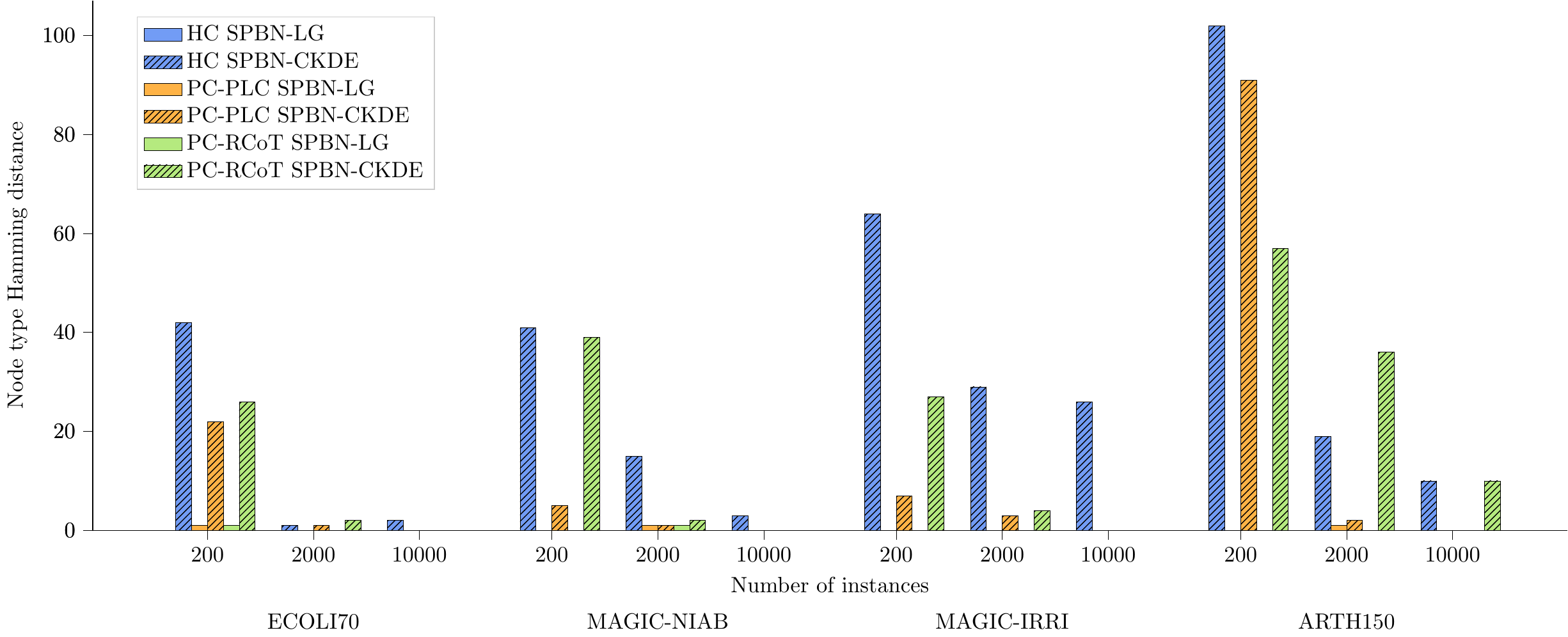}
\end{center}
\caption{THMD of SPBNs to the ground truth model.}\label{fig:gbn_thd}
\end{figure}

We show the THMD value between the SPBN models and the true model in Figure~\ref{fig:gbn_thd}. In this experimental framework, THMD was equal to the number of CKDE nodes, as all nodes in the true model were of the LG type. It is clear that SPBN-LG outperformed SPBN-CKDE in finding the best node types. This is reasonable because the starting point of SPBN-LG is optimal in the node types, while the SPBN-CKDE is the worst model possible in the search space. However, there is a clear trend towards a THMD reduction when a larger sample size is available. For small sample sizes, it is possible that a normality test such as Shapiro-Wilks (as suggested in Section~\ref{sec:pc_algorithm}) could reduce the THMD, so we leave that analysis as future work.

\begin{figure}
\begin{center}
\includegraphics[width=\linewidth]{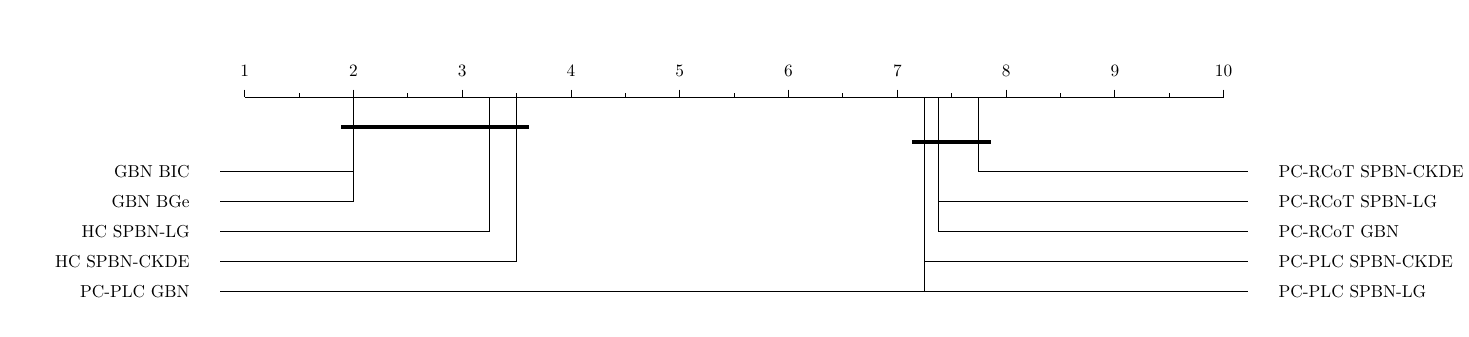}
\end{center}
\caption{Critical difference diagram for the mean rank of each algorithm trained using the data sampled from GBNs.}\label{fig:cd_gaussian_diagram}
\end{figure}

The log-likelihood of the test dataset in the trained models is shown in Table~\ref{tab:gbn_loglikelihood}. We observe rather similar results for all models and datasets when using the same learning algorithm. The HC algorithm tends to have a higher log-likelihood than the PC algorithm, which is coherent because the HC search process is guided by the improvement of the log-likelihood. To detect statistically significant differences between all algorithms, we performed a Friedman test with $\alpha = 0.05$ and a Bergmann-Hommel post-hoc procedure to detect the pairwise significant differences~\cite{CompareAlgorithms}. We illustrate the obtained results graphically in Figure~\ref{fig:cd_gaussian_diagram} using a critical difference diagram~\cite{Demsar} that represents the mean rank of each algorithm. The horizontal black lines connect the groups of algorithms that do not have a significant difference. The models learned with the HC algorithm have a statistical significant difference with the models learned with PC. However, the differences are not statistically significant between learning GBNs or SPBNs using the same algorithm. Therefore, we concluded that SPBN learning was as suitable as GBN learning for the training data sampled from a GBN.

{
\newlength{\collength}
\setlength{\collength}{1.44cm}
\setlength{\tabcolsep}{1pt}
\begin{table}
\caption{Log-likelihood of the test dataset in the trained models using the data sampled from GBNs. We also show the log-likelihood of the test dataset in the ground truth model for reference (the best result for each dataset is highlighted boldface).}
\label{tab:gbn_loglikelihood}
\scriptsize
\begin{center}
\begin{tabular}{L{3cm}C{\collength}C{\collength}C{\collength}C{\collength}C{\collength}C{\collength}}
\hline
GBN&\multicolumn{3}{c}{ECOLI70}&\multicolumn{3}{c}{MAGIC-NIAB}\\
Instances&200&2,000&10,000&200&2,000&10,000\\
\hline
True model&\multicolumn{3}{c}{$-41522.34$}&\multicolumn{3}{c}{$-48469.84$}\\
GBN BIC&$-42426.52$&$-41610.22$&$-41529.44$&$-49682.02$&$-48537.96$&$\mathbf{-48476.6}$\\
GBN BGe&$-42372.28$&$-41592.76$&$\mathbf{-41526.69}$&$\mathbf{-49587.08}$&$\mathbf{-48536.56}$&$-48479.21$\\
HC SPBN-LG&$\mathbf{-42258.08}$&$\mathbf{-41580.96}$&$-41528.95$&$-49712.7$&$-48577.6$&$-48497.42$\\
HC SPBN-CKDE&$-45033.49$&$-41638.3$&$-41529.25$&$-50365.1$&$-48642.04$&$-48486.65$\\
PC-PLC GBN&$-44759.66$&$-44286.85$&$-43274.39$&$-49637.17$&$-48572.77$&$-48488.73$\\
PC-RCoT GBN&$-47721.66$&$-43604.97$&$-43013.12$&$-49655.38$&$-48678.99$&$-48513.3$\\
PC-PLC SPBN-LG&$-44818.33$&$-44286.85$&$-43274.39$&$-49637.17$&$-48577.39$&$-48488.73$\\
PC-RCoT SPBN-LG&$-47780.34$&$-43604.97$&$-43013.12$&$-49655.38$&$-48683.61$&$-48513.3$\\
PC-PLC SPBN-CKDE&$-45406.38$&$-44294.19$&$-43274.39$&$-49778.77$&$-48577.39$&$-48488.73$\\
PC-RCoT SPBN-CKDE&$-48381.11$&$-43618.0$&$-43013.12$&$-50333.59$&$-48684.85$&$-48513.3$\\\\\hline
GBN&\multicolumn{3}{c}{MAGIC-IRRI}&\multicolumn{3}{c}{ARTH150}\\
Instances&200&2,000&10,000&200&2,000&10,000\\
\hline
True model&\multicolumn{3}{c}{$-76193.11$}&\multicolumn{3}{c}{$-36471.74$}\\
GBN BIC&$-78312.61$&$\mathbf{-76322.27}$&$\mathbf{-76209.87}$&$-41745.83$&$\mathbf{-36709.18}$&$-36495.68$\\
GBN BGe&$-77986.66$&$-76353.57$&$-76213.31$&$-43537.06$&$-36755.5$&$-36500.39$\\
HC SPBN-LG&$\mathbf{-77638.45}$&$-76377.09$&$-76218.89$&$\mathbf{-38841.11}$&$-36709.64$&$\mathbf{-36484.31}$\\
HC SPBN-CKDE&$-79465.69$&$-76576.7$&$-76279.79$&$-43112.32$&$-36837.14$&$-36510.63$\\
PC-PLC GBN&$-78319.14$&$-76844.8$&$-76598.0$&$-42207.35$&$-39526.29$&$-38491.47$\\
PC-RCoT GBN&$-78340.23$&$-76653.77$&$-76356.27$&$-45827.92$&$-41233.34$&$-40556.53$\\
PC-PLC SPBN-LG&$-78319.14$&$-76844.8$&$-76598.0$&$-42207.35$&$-39532.54$&$-38491.47$\\
PC-RCoT SPBN-LG&$-78340.23$&$-76653.77$&$-76356.27$&$-45827.92$&$-41233.34$&$-40556.53$\\
PC-PLC SPBN-CKDE&$-78517.12$&$-76857.15$&$-76598.0$&$-44519.13$&$-39539.88$&$-38491.47$\\
PC-RCoT SPBN-CKDE&$-78751.04$&$-76672.95$&$-76356.27$&$-46582.94$&$-41370.77$&$-40564.74$\\\hline
\end{tabular}
\end{center}
\end{table}
}

\subsection{UCI Repository Data}

In this section, we present the results of testing the SPBN model learning based on the real data extracted from the UCI repository. In this experimental framework, we did not have the information about the structure of an underlying Bayesian network that produced the data. Table~\ref{tab:uci_datasets} presents the number of instances ($N$) and variables ($n$) in each dataset. Some of these datasets were designed for classification so that they included a discrete class variable. Every class/discrete variable was removed from each dataset. The information shown in Table~\ref{tab:uci_datasets} reflects the number of variables after the removal procedure.

We compared the different types of continuous Bayesian networks: KDEBNs, GBNs, and SPBNs. We did not include discrete Bayesian networks because they model a probability mass function, as opposed to the continuous Bayesian networks that model probability density functions. As we did not know the structure of the underlying Bayesian network (if any) that produced the data, we tested the density estimation capabilities of each model. For this purpose, we applied a 10-fold cross-validation approach to estimate the log-likelihood of the unseen data in each model. Accordingly, ten models were trained using different training folds, and the log-likelihood was estimated in the unseen instances of the test fold. The estimation of the expected log-likelihood in the unseen data was derived as the mean test log-likelihood for every fold. For HC, the KDEBNs were learned using a procedure similar to the SPBN learning procedure described in Algorithm~\ref{alg:greedy_hill_climbing} with a cross-validation score and a validation dataset to detect convergence. However, the \textsc{Type-Change} operator is not valid to learn the structure of a KDEBN model (i.e. its node type is fixed in advance). We tested the same learning configurations for GBNs and SPBNs as in the previous sections. The selection of $\lambda = 5$ often returned better models than $\lambda = 0$ (for both KDEBNs and SPBNs), so we present only the results with $\lambda = 5$ in this section.

\begin{figure}
\begin{center}
\includegraphics[width=\linewidth]{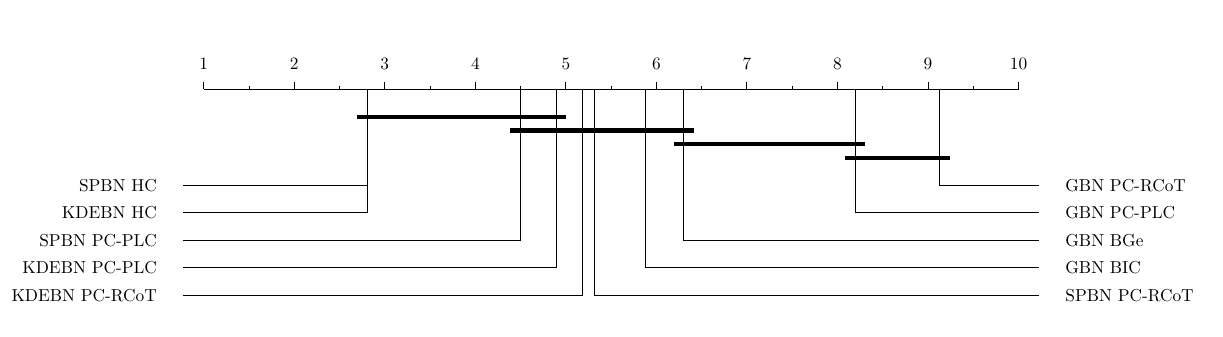}
\end{center}
\caption{Critical difference diagram for the mean rank of each algorithm in the UCI datasets.}\label{fig:cd_diagram}
\end{figure}

As in the previous section, we performed a Friedman test with $\alpha = 0.05$ and a Bergmann-Hommel post-hoc procedure. The critical difference diagram is represented in Figure~\ref{fig:cd_diagram}. We concluded that SPBNs and KDEBNs perform better than GBNs because their differences in the expected log-likelihood were statistically significant. This was reasonable because, in general, the real data do not follow the multivariate Gaussian distribution. Also, the models trained with HC are the top ranked, except GBN BIC and GBN BGe. This is meaningful because the optimization criterion of HC is the cross-validated log-likelihood unlike the conditional independence tests in PC. The partial linear correlation test tends to have a better mean rank than RCoT, even though there is no statistically significant difference. This suggests that there are some linear relationships between the variables in the real data, so the partial correlation test is competitive with respect to RCoT. In addition, since the GBNs show a significant lower log-likelihood than the more flexible models, the real data probably also contain nonlinear relationships. This emphasizes the importance of modeling explicitly a combination of linear and nonlinear relationships as SPBNs do.

\begin{figure}
\begin{center}
\includegraphics[width=\linewidth]{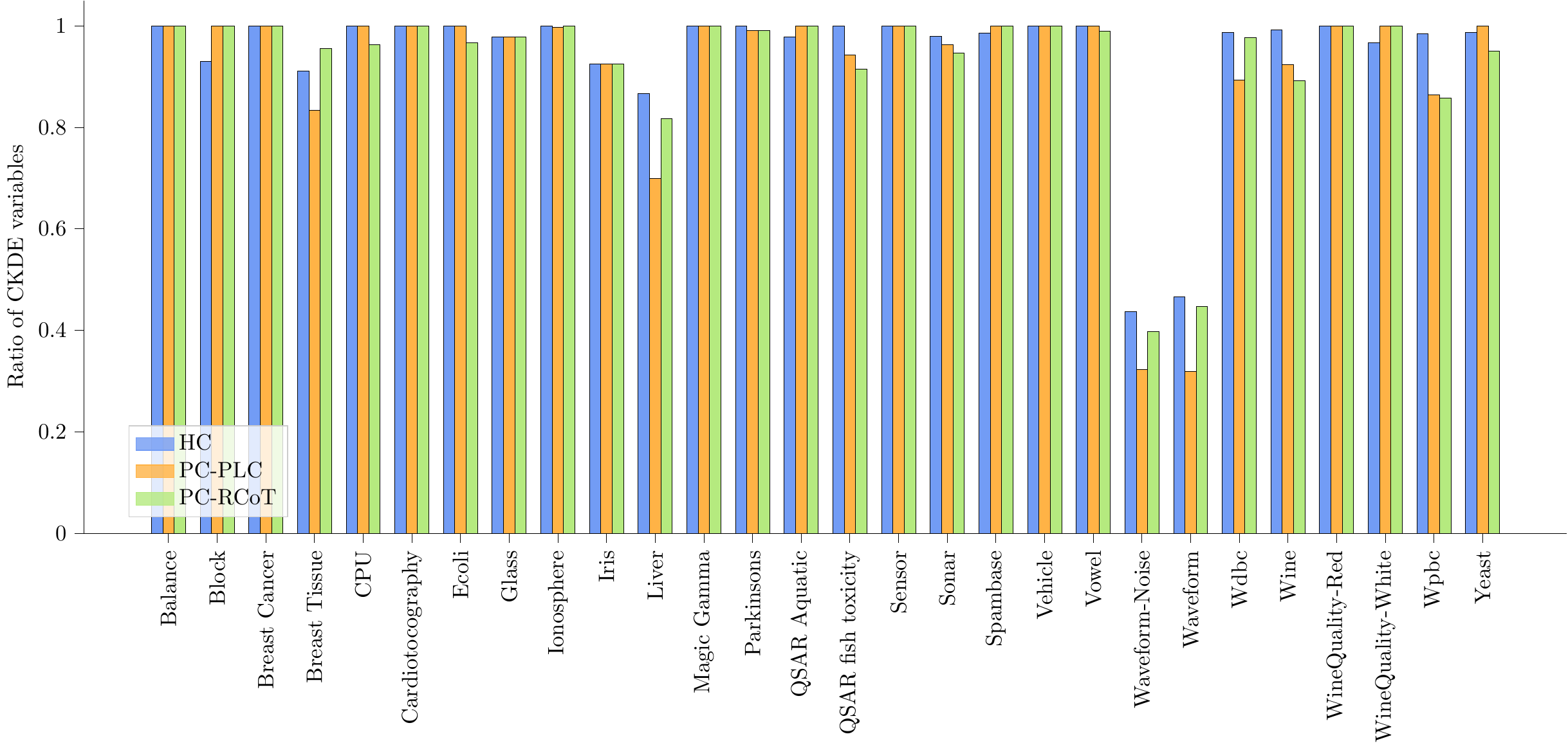}
\end{center}
\caption{Ratio of CKDE nodes on different datasets learned with HC (blue), PC-PLC (orange) and PC-RCoT (green) algorithms.}\label{fig:kdeness}
\end{figure}

The KDEBN and SPBN models obtain a fairly similar mean rank for all learning configurations, so there is not statistically significant difference. In Figure~\ref{fig:kdeness}, we represent the ratio of CKDE nodes in the SPBN models for all the learning algorithms. The three different algorithms return a similar number of CKDE nodes. We can see that the proportion of CKDE nodes is quite high, and for about half of the datasets, every node type was CKDE; and therefore, according to Proposition~\ref{prop:kde_networks}, the final SPBN was equivalent to a KDEBN network. This justifies why there was no statistically significant difference between the SPBN and KDEBN models: most datasets were better represented by KDEBN models or the models with many CKDE CPDs. Noticeably, Waveform and Waveform-Noise datasets have the lowest proportion of CKDE nodes. This is explained by the fact that these datasets include variables with Gaussian noise, which the SPBN model was able to detect correctly.

\begin{table}
\caption{Datasets from the UCI repository.}
\label{tab:uci_datasets}
\begin{center}
\begin{tabular}{lll@{\hskip 0.4in}lll}
\hline
Dataset & $N$ & $n$ & Dataset & $N$ & $n$\\
\hline
Balance & 625 & 4 & QSAR fish toxicity & 908 & 7\\
Block & 5473 & 10 & Sensor & 5456 & 24\\
Breast Cancer & 683 & 9 & Sonar & 208 & 60\\
Breast Tissue & 106 & 9 & Spambase & 4601 & 57\\
CPU & 209 & 8 & Vehicle & 846 & 18\\
Cardiotocography & 2126 & 19 & Vowel & 990 & 10\\
Ecoli & 336 & 5 & Waveform & 5,000 & 21\\
Glass & 214 & 9 & Waveform-Noise & 5,000 & 40\\
Ionosphere & 351 & 33 & Wdbc & 569 & 30\\
Iris & 150 & 4 & Wine & 178 & 13\\
Liver & 345 & 6 & WineQuality-Red & 1599 & 12\\
Magic Gamma & 19020 & 10 & WineQuality-White & 4898 & 12\\
Parkinsons & 195 & 21 & Wpbc & 194 & 33\\
QSAR Aquatic & 546 & 9  & Yeast & 1484 & 8\\
\hline
\end{tabular}
\end{center}
\end{table}

\subsection{Monitoring Bearings Degradation}

In this section, we use SPBNs to monitor the degradation of rolling bearings. Rolling bearings are one of the most commonly used elements in industrial machines. Usually, these bearings suffer from degradation and can be a cause of machine breakdowns. For this reason, monitoring the state of bearing degradation can be a useful technique in machine maintenance. We will use the data provided by PRONOSTIA~\cite{PRONOSTIA}, which is an experimentation platform that degrades the bearings in a few hours. The data is captured with two accelerometers in the horizontal and vertical axes, to detect the bearing vibration. As is common in bearing diagnostics, we will analyze the data in the frequency domain, focusing on some frequencies of interest and their harmonics: ball pass frequency outer race (BPFO), ball pass frequency inner race (BPFI), fundamental train frequency (FTF) and ball spin frequency (BSF)~\cite{BearingsTutorial}.

\begin{figure}
	\begin{center}
	\includegraphics[width=\linewidth]{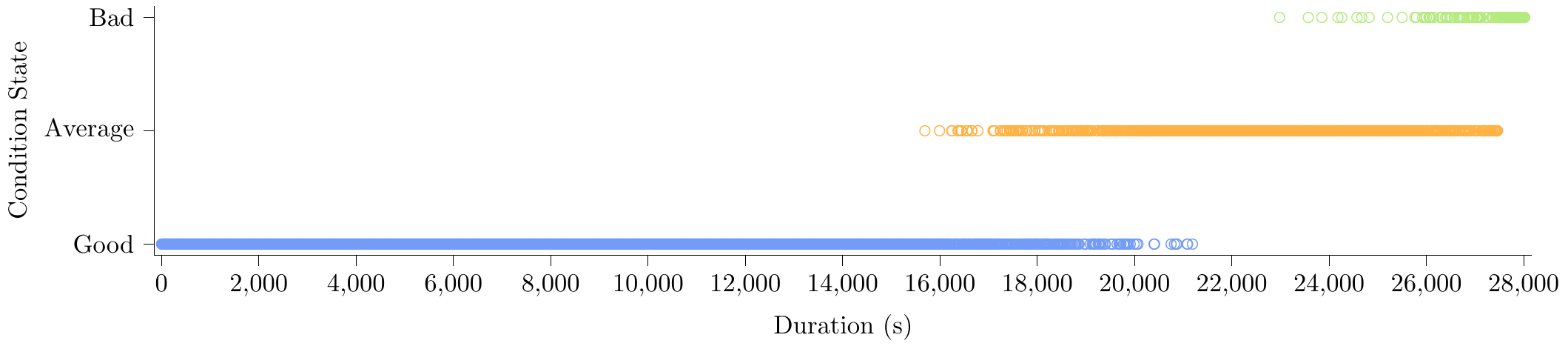}
	\end{center}
	\caption{Segmentation of a bearing dataset (Bearing1\_1) into good, average and bad state instances.}\label{fig:bearing_segmentation}
\end{figure}

The PRONOSTIA dataset provides data with three different load conditions, but in this section, we will only use the first load condition. The training dataset contains two different bearings that were run to failure. To construct a model of the normal behaviour of a bearing, we segmented the data into three different state conditions: good state, average state and bad state. We detected these condition states using a hidden Markov model (HMM) assuming Gaussian emissions~\cite{HMMs}. Typically, the good state is at the start of the data and exhibits low amplitudes for the frequencies studied. The average and bad states are usually located at the middle and the end of the data, and have average and high amplitudes respectively. Figure~\ref{fig:bearing_segmentation} shows the segmentation found for a training dataset into good, average and bad state. From this segmentation, we can learn three different SPBNs to model the good, average and bad state using the two learning training datasets.

\begin{figure}
	\begin{center}
	\includegraphics[width=\linewidth]{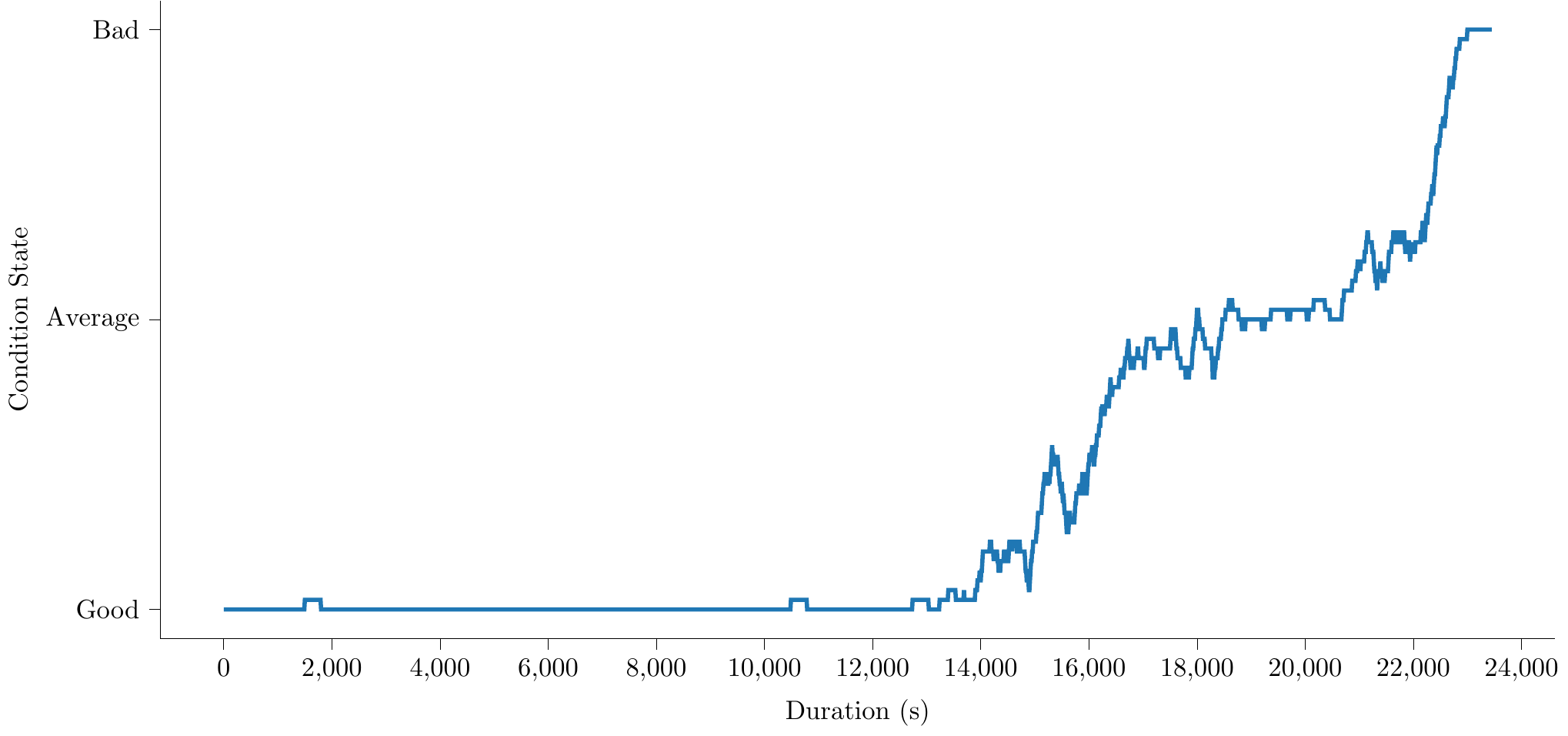}
	\end{center}
	\caption{Estimated degradation process of a bearing dataset (Bearing1\_3).}\label{fig:bearing_process}
\end{figure}

Learning three models of the bearing state can help us to track the degradation process of the bearing. For each instance of a test dataset, we can detect which SPBN model provides the larger log-likelihood. Figure~\ref{fig:bearing_process} shows a gradual degradation process for a given bearing, by selecting the best model on each instance and applying a moving average to smooth the final result. Other bearings show different degradation patterns, e.g. abrupt degradation.

\begin{figure}
	\begin{center}
	\includegraphics[width=\linewidth]{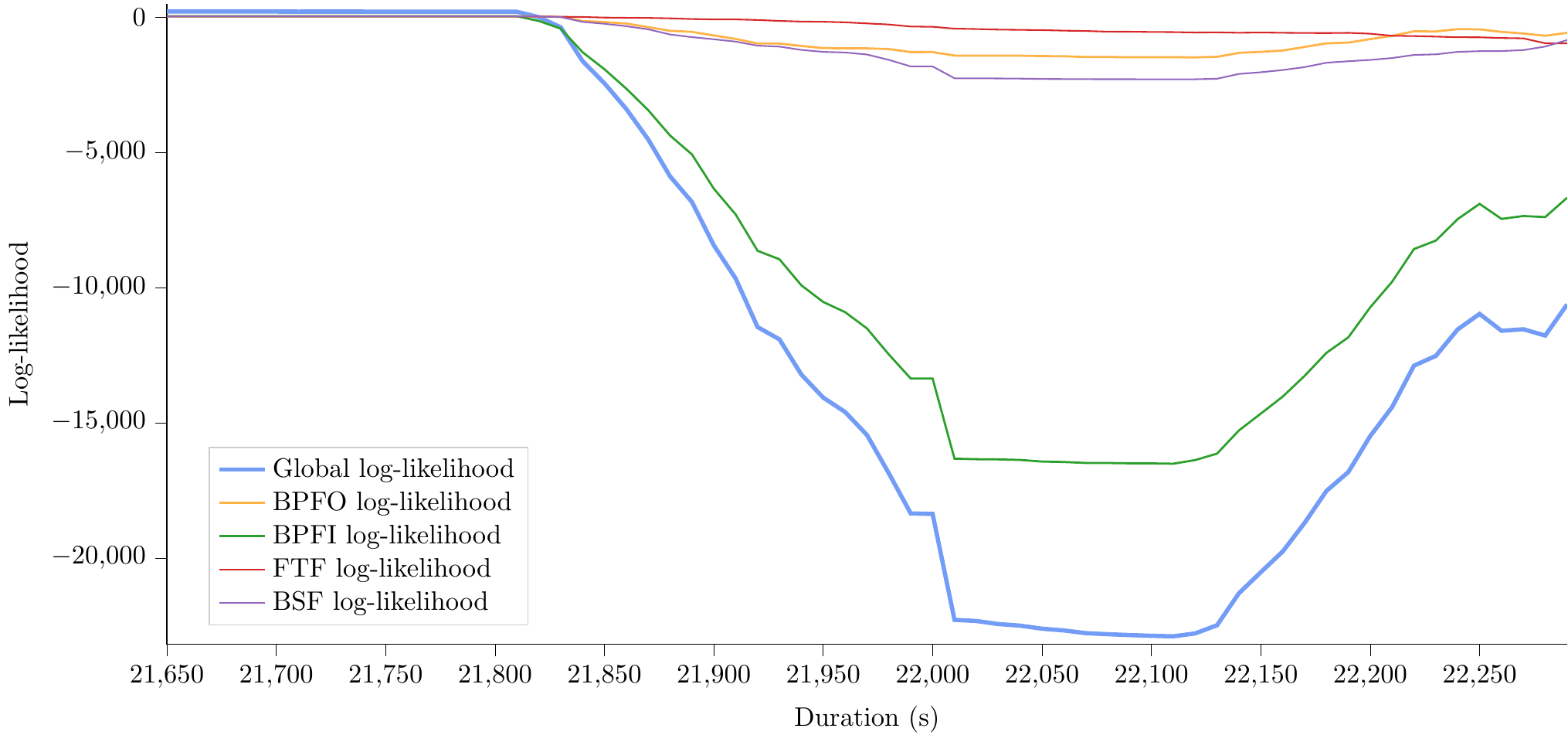}
	\end{center}
	\caption{Global log-likelihood and local log-likelihood for the BPFO, BPFI, FTF and BSF frequencies and their harmonics of a bearing dataset (Bearing1\_7) according to the good state model.}\label{fig:bearing_local_defect}
\end{figure}

One of the main features of Bayesian networks is their interpretability because the contribution of each variable to the global log-likelihood can be analyzed. Figure~\ref{fig:bearing_local_defect} shows a decrease in the log-likelihood of a test bearing dataset according to the good state model. Also, we included the local log-likelihood contribution of each frequency and their harmonics. We can see that this decrease in the log-likelihood is mainly explained by abnormal BPFI amplitudes. This information suggests that a defect has ocurred in the inner race of the bearing.

This work has a clear temporal component. Therefore, an interesting alternative would be the use of dynamic Bayesian networks. We leave as a future research line the creation of dynamic semiparametric Bayesian networks.

\subsection{Execution Times}\label{sec:execution_times}

In this section, we show how the learning time of each learning procedure compares in practice. We created three synthetic models with different number of variables: a small model with 5 variables (the model in Figure~\ref{fig:true_model}), a medium-size model with 10 variables and a large model with 20 variables (not shown). We defined half of the nodes with a linear Gaussian relationship and the other half using nonlinear relationships. For each model we sampled datasets with different number of instances: 200, 500, 2000, 4000 and 10000 instances. Then, we measured the execution time of all learning algorithms by repeating the learning process several times (with different sets of indices $\mathcal{I}$) and calculating the average value. For HC, we set $\lambda = 0$.

\begin{figure}
	\begin{center}
	\includegraphics[width=\linewidth]{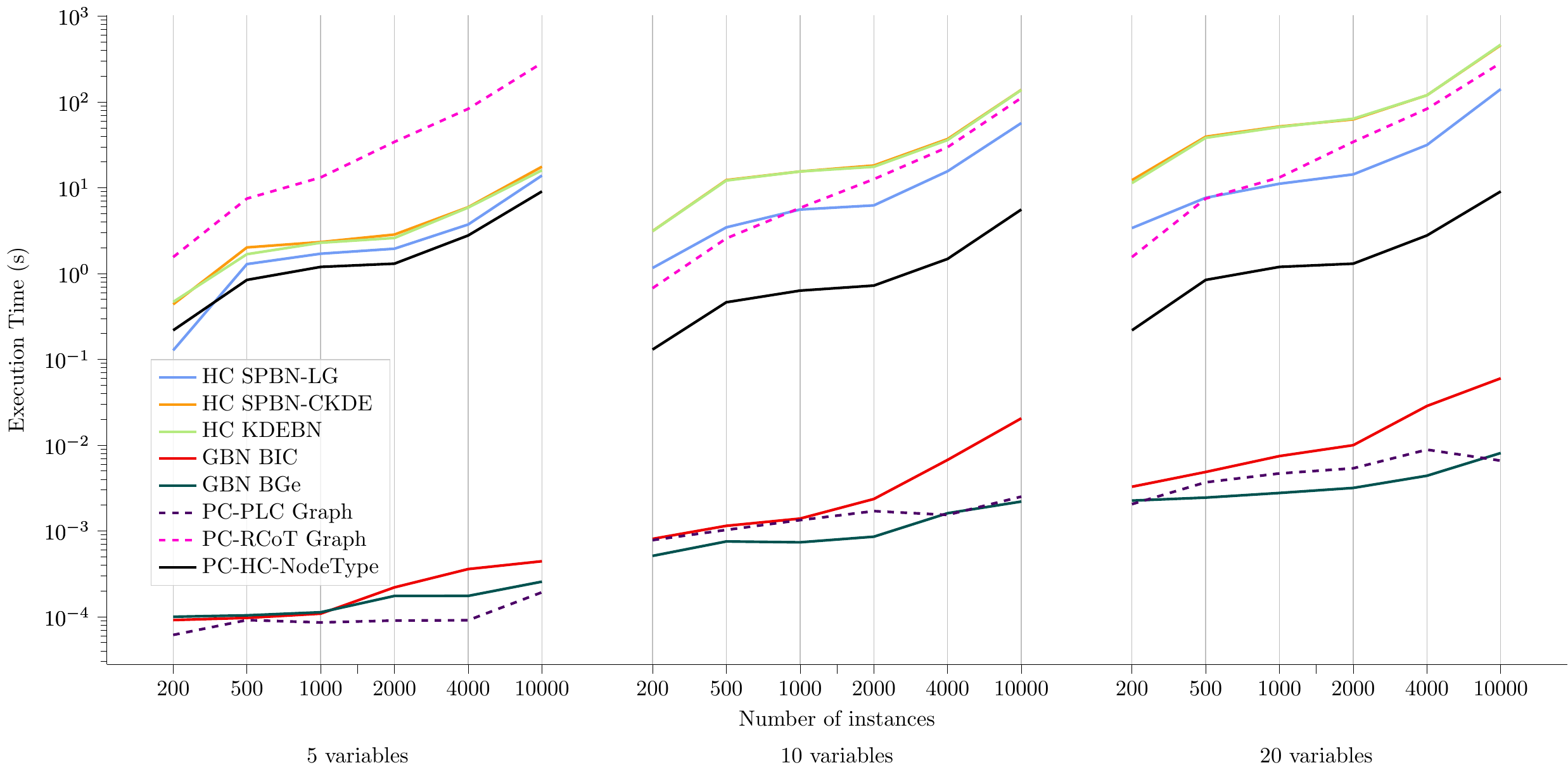}
	\end{center}
	\caption{Execution times of all the learning procedures with different number of training instances and variables.}\label{fig:execution_times}
\end{figure}

Figure~\ref{fig:execution_times} presents the execution time for all the learning algorithms in logarithm scale. The HC and PC algorithm times are shown with solid and dashed lines respectively. The PC-PLC Graph and PC-RCoT Graph show the time to learn just the graph of the model. This is the final model for the GBN and KDEBN models. To finish the learning process of a SPBN, an HC is executed that selects the best node types (PC-HC-NodeType). From these results we can see that the most performant methods are the HC with the BIC and BGe scores, and the PC-PLC algorithm. All of these learning procedures make parametric assumptions. Recall also that in BGe and PLC, some information can be cached at the start of the algorithm so the asymptotic time complexity only depends on the number of variables during the HC and PC iterations. Furthermore, we can check that the PC-HC-NodeType algorithm is usually much faster than the HC SPBN algorithm that also has to search for the best set of arcs. In all runs, HC SPBN-LG was faster than HC KDEBN. This is because the initial delta score cache for SPBN-LG takes less time. We discussed the difference in the asymptotic complexity of the cross-validated score (Equation~\ref{eq:cv_loglikelihood}) between the LG and CKDE CPDs in Section~\ref{sec:time_complexity}. This can be easily verified by seeing that the HC SPBN-CKDE and HC KDEBN models take almost the same time in all runs.

\section{Conclusion}\label{sec:conclusion}

In the present paper we introduced a new class of Bayesian networks called semiparametric Bayesian networks that could be applied to model continuous data using both parametric and nonparametric estimation models. The class of SPBNs includes every possible GBN and every possible KDEBN. GBNs are fully parametric models, and KDEBNs correspond to fully nonparametric ones. In between these two extreme cases, an SPBN framework could be used to build a network in which some parts were parametric while other parts were nonparametric. Therefore, this approach allowed automatically adopting the advantage of parametric assumptions when appropriate, while also providing the flexibility of nonparametric models when necessary.

We proposed learning SPBNs based on the HC and PC algorithms. We note that other state-of-the-art learning algorithms could also be considered. Notably, as the proposal introduced in this paper is general, other score and search algorithms can be implemented using the \textsc{Type-Change} operator (Section~\ref{sec:learning_structure}).

The results of the conducted experiments indicated that SPBNs could be implemented finding a suitable combination of parametric and nonparametric components. If the considered data followed the Gaussian distribution, the corresponding learned SPBN tends to be a GBN model. However, if the data clearly did not belong to the Gaussian distribution, the proposed learning algorithm produced more flexible SPBNs that combined the advantages of GBNs and KDEBNs.

There are multiple research lines that can be further investigated in the future. A better bandwidth selection for CKDE could improve the density estimation results. Discrete variables can be included to develop hybrid SPBNs. In addition, introducing a tractable inference algorithm to perform queries in SPBNs would be of great interest. Moreover, temporal data could be better analyzed using dynamic semiparametric Bayesian networks. To conclude, we also plan to train SPBN classifiers in the future.

\section*{Acknowledgments}

This work has been partially supported by the Ministry of Education, Culture and Sport through the grant FPU16/00921 and by the Spanish Ministry of Science and Innovation through the PID2019-109247GB-I00 and RTC2019-006871-7 projects.

\bibliography{mybibfile}

\end{document}